\pdfoutput=1

\documentclass[11pt]{article}

\usepackage[preprint]{acl}

\usepackage{times}
\usepackage{latexsym}

\usepackage[T1]{fontenc}

\usepackage[utf8]{inputenc}

\usepackage{microtype}

\usepackage{inconsolata}

\usepackage{tabularx}  
\usepackage[normalem]{ulem}  
\usepackage{array}  
\usepackage[absolute,overlay]{textpos}
\usepackage{graphicx}
\usepackage{lipsum} 
\usepackage{amsmath, amssymb}
\usepackage{mathtools}
\usepackage[skip=2pt]{caption} 
\usepackage{subcaption}  
\usepackage{booktabs}
\usepackage{multirow}
\usepackage{enumitem}
\usepackage{float}
\usepackage{makecell}
\usepackage{authblk}

\usepackage{listings}  
\usepackage{tcolorbox} 

\newtcolorbox{mybox}[1][]{
  colback=white,       
  colframe=black,      
  coltitle=white,      
  title={#1},          
  fonttitle=\bfseries, 
  sharp corners,       
  boxrule=0.5mm        
}
\newcolumntype{L}{>{\raggedright\arraybackslash}X} 

\DeclareUnicodeCharacter{2033}{''}

\title{Are Emotions Arranged in a Circle? Geometric Analysis of Emotion Representations via Hyperspherical Contrastive Learning}

\author{
    \textbf{Yusuke Yamauchi}\textsuperscript{1} \quad \textbf{Akiko Aizawa}\textsuperscript{2,1} \\
    \textsuperscript{1}The University of Tokyo, 
    \textsuperscript{2}National Institute of Informatics \\
    \texttt{\{y\_yamauchi@is.s.u-tokyo.ac.jp, aizawa@nii.ac.jp\}}
}

\begin{document}
\maketitle

\begin{abstract}
Psychological research has long utilized circumplex models to structure emotions, placing similar emotions adjacently and opposing ones diagonally. Although frequently used to interpret deep learning representations, these models are rarely directly incorporated into the representation learning of language models, leaving their geometric validity unexplored.
This paper proposes a method to induce circular emotion representations within language model embeddings via contrastive learning on a hypersphere. We show that while this circular alignment offers superior interpretability and robustness against dimensionality reduction, it underperforms compared to conventional designs in high-dimensional settings and fine-grained classification. Our findings elucidate the trade-offs involved in applying psychological circumplex models to deep learning architectures. Our code is available at \url{https://github.com/yama11235/EmpiricalCircumplexModel}
\label{sec:abstract}
\end{abstract}

\section{Introduction}
In recent years, the mechanistic interpretability of large language models (LLMs) has emerged as a pivotal field for ensuring AI safety and controllability~\cite{zhao2024explainability, bereska2024mechanistic, opitz-etal-2025-interpretable}. Central to this field are the Linear Representation Hypothesis and Superposition Hypothesis, which posit that models represent human-interpretable concepts as linear directions within low-dimensional subspaces~\cite{pmlr-v235-park24c, elhage2022toymodelssuperposition}. This framework has been empirically validated through representation engineering, where intervening in specific directions allows for the direct manipulation of model behavior~\cite{li2023inference, Arditi2024RefusalIL}.

\begin{figure}
    \centering
    \includegraphics[width=\linewidth]{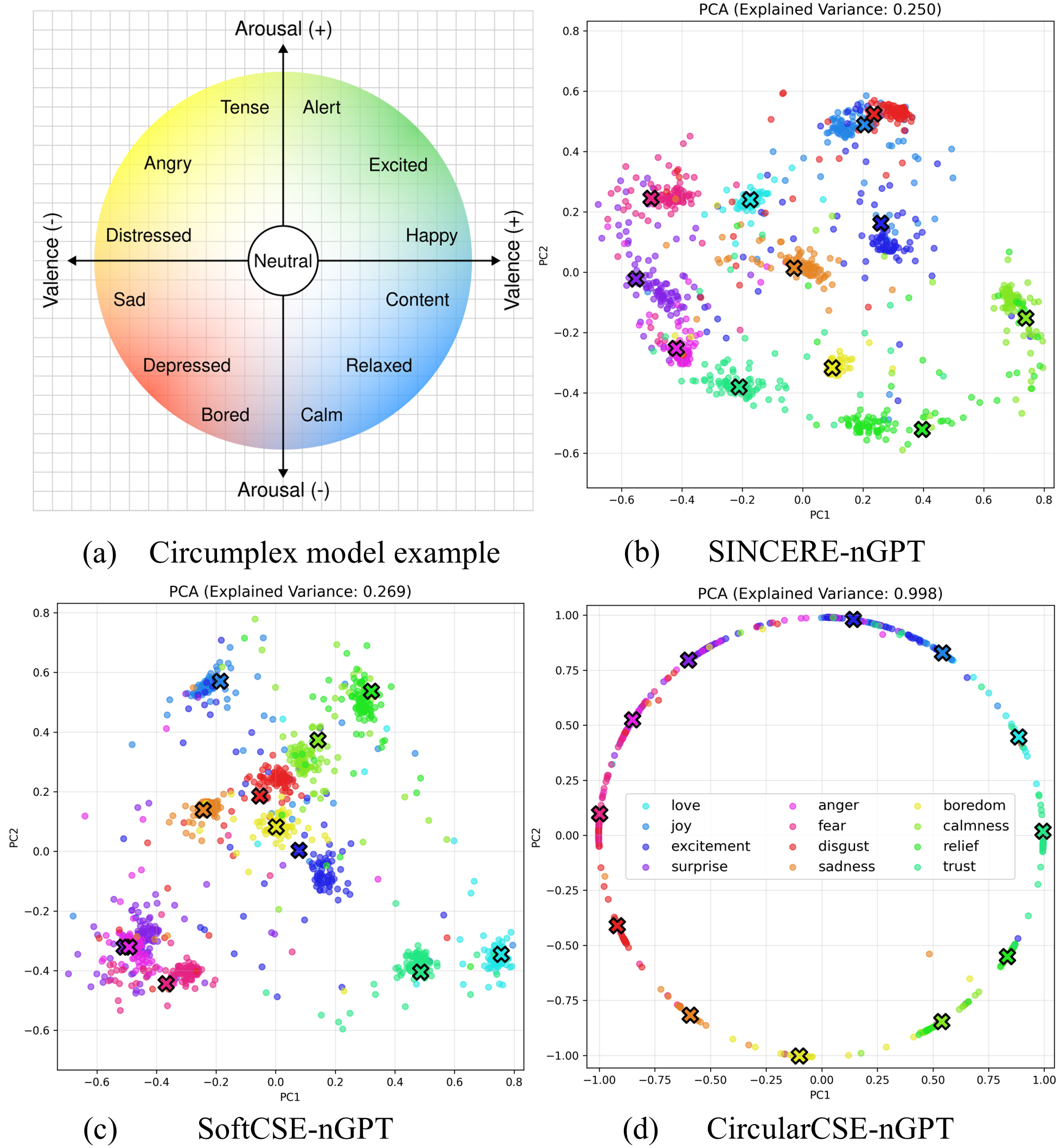}
    \caption[An example of the psychological Circumplex Model of Affect]{(a) An example of the psychological circumplex model of emotion\protect\footnotemark. (b)(c)(d) PCA plots of the embeddings from the models trained in this study.}
    \label{fig:pca2_visualization}
\end{figure}

\footnotetext{Based on the circumplex model by Russell (1980). Retrieved from Wikimedia Commons (\url{https://en.wikipedia.org/wiki/File:Circumplex_model_of_emotion.svg}).}

However, recent studies suggest that not all concepts are best captured by linear structures. Periodic concepts, such as days of the week or months of the year, have been found to form circular embeddings, and models undergoing the "grokking" phenomenon in modular arithmetic ultimately arrange numerical representations in circular configurations~\cite{engels2025not, park2025iclr,liu2022towards, nanda2023progress}. These findings raise a fundamental question: Are there other critical human concepts that are characterized by non-linear, specifically circular, manifold structures?

In psychology, the Circumplex Model of Affect has long proposed that emotions are arranged in a circle defined by two axes: valence and arousal (Figure~\ref{fig:pca2_visualization}(a)). Psychological emotion models are frequently employed in the analysis of machine learning models, with research investigating whether language models actually reflect these underlying structures~\cite{10902477, zhao2025emergencehierarchicalemotionorganization, reichman2025emotions}. However, since these analyses are often limited to post-hoc observations, such as centroid distances or label co-occurrence probabilities, they remain confined to analyzing general trends and do not extend to an analysis at the concept manifold level.
Whether a language model expresses emotions in the same way humans do, or whether it should do so, cannot be rigorously discussed unless the manifold structure of the model’s embedding representations is explicitly reconstructed.

This paper investigates the validity of the circular emotion structure by explicitly inducing it within the embedding space. To this end, we first collect and synthesize a text dataset annotated with emotions assumed to follow a circular structure. We then train emotion representations using nGPT~\cite{loshchilov2025ngpt}, an architecture designed for hyperspherical representation learning, in combination with three contrastive loss functions: SINCERE~\cite{feeney2023sincere}, SoftCSE~\cite{zhuang2024not}, and our proposed CircularCSE. These objectives calibrate distances between emotion labels according to psychologically grounded circular distances (CD)~\cite{zhao2024err}. Finally, we evaluate the resulting representations across three backbone categories (BERT-like models, LLM-based encoders, and decoder-only LLMs) using a comprehensive set of metrics, including V-Measure for discriminative power and CD-r for psychological alignment. 

Our results reveal a stark structural dilemma. While the circular structure (CircularCSE) offers superior interpretability and remains robust in low-dimensional spaces or with few labels (Figure~\ref{fig:pca2_visualization}(d)), its performance degrades significantly in high-dimensional or fine-grained label settings compared to conventional designs. We provide a theoretical explanation for this: SINCERE thrives in high dimensions by arranging labels as an orthogonal simplex ($90^{\circ}$ margins), whereas CircularCSE's 2D ring geometry forces much tighter boundary margins, inherently limiting discriminability as the number of labels increases.

This conflict suggests that aligning models with human interpretation, which implicitly assumes low-dimensional manifold structures, comes at a structural cost to discriminative power. However, aligning model representations with human psychology offers diverse benefits, ranging from intuitive mechanistic interpretation to reduced computational costs. Our research re-examines the validity of incorporating human interpretability into model design from the perspective of deep learning.

\begin{figure*}
    \centering
    \includegraphics[width=\linewidth]{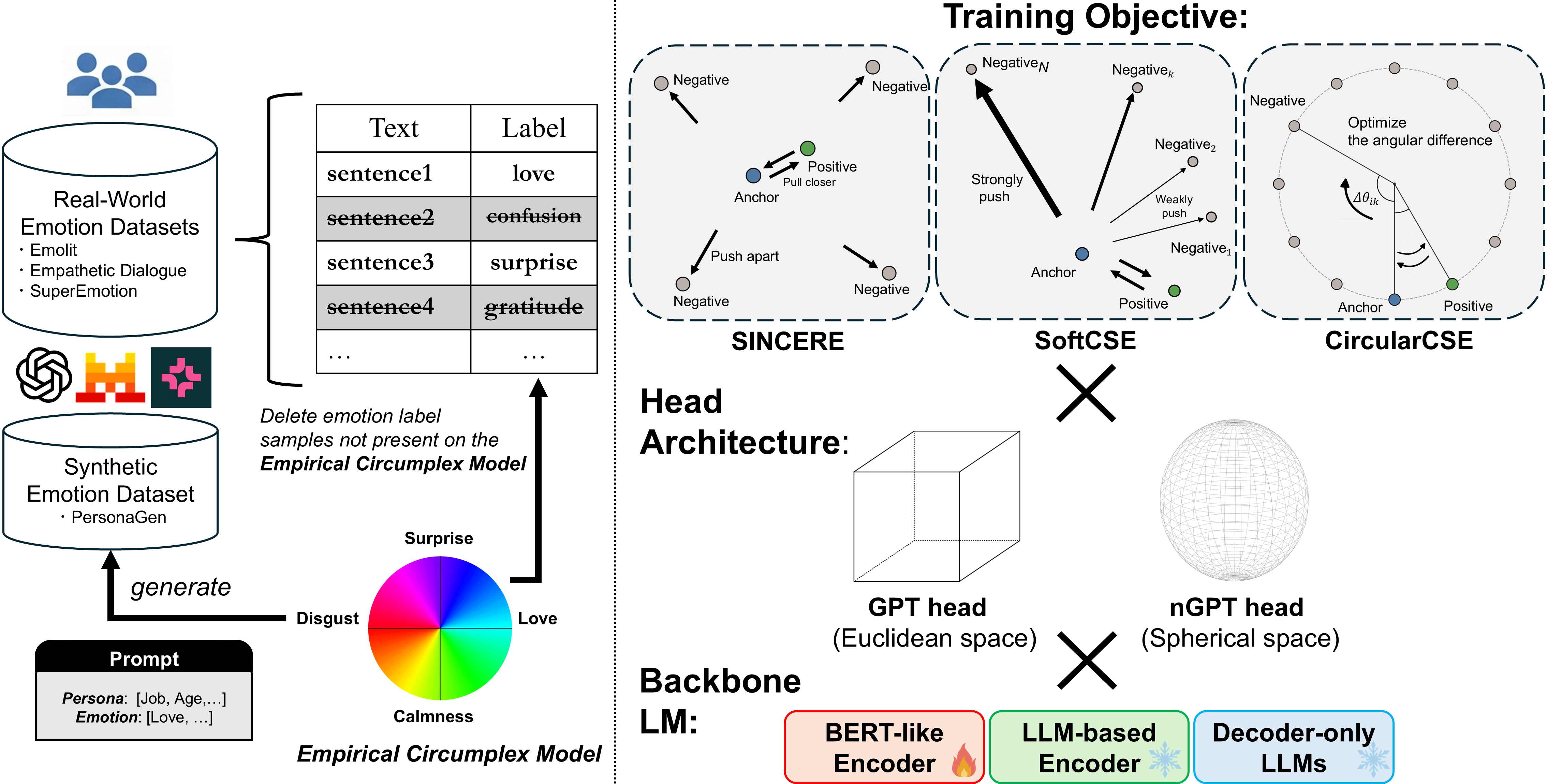}
    \caption{\textbf{Overview of our experimental framework.} (Left) Dataset construction procedure. Corresponding emotion labels are extracted or synthesized to reproduce the circumplex emotion structure. (Right) Training of GPT or nGPT heads across three backbone architectures using three distinct loss functions.}
    \label{fig:Overview_framework}
\end{figure*}
\section{Related Work}
\textbf{Psychological models of emotion} are generally classified based on whether they represent emotion in a continuous or discrete space. A representative example of a continuous model is Russell’s Circumplex Model of Affect, which posits that emotions are represented along two axes: valence and arousal~\cite{Russell1980}. Along with the PAD model, which adds a dominance axis, it has been widely adopted in recent years~\cite{Mehrabian1974AnAT}. In contrast, discrete emotion models construct a space from a set of basic emotions, where diverse emotional states are expressed through the compounding or subdivision of these basics. Plutchik proposed the "Wheel of Emotions," consisting of eight basic emotions~\cite{Plutchik1980Emotion}. Regardless of the continuous or discrete space, most foundational psychological models represent emotions as a circular structure~\cite{shaver1987emotion, ekman1992argument}. This design reflects the clear similarities (e.g., joy and excitement) and polarities (e.g., positive and negative) inherent in emotions. 

\noindent\textbf{Geometry of LLM Representations.}
While the analysis of emotion representations in LLMs is common, many studies focus on examining the macroscopic relationships between emotions in existing models based on co-occurrence probabilities or the Euclidean distances between centroids~\cite{guo2021enhancing, 10902477, zhao2025emergencehierarchicalemotionorganization, reichman2025emotions}. Outside the domain of emotion, it has been discovered that periodic concepts and numerical values in modular addition tasks exhibit circular structures~\cite{engels2025not, park2025iclr,liu2022towards, nanda2023progress}. However, these findings are limited to specific conditions, such as those involving sparse autoencoders~\cite{bricken2023monosemanticity} or grokking. To the best of our knowledge, no study has successfully constructed a circular structure for emotion representations within a standard language model.

\noindent\textbf{Design of Contrastive Learning.} Although many studies employ improved or custom variants of contrastive loss functions, they are primarily designed to enhance discriminative performance, with few focusing on the explicit induction of manifold structures~\cite{choi2020amclossangularmargincontrastive, yang2021circular, Deng_2022}. In the domain of image emotion classification, a prior study has incorporated angular differences based on a circular assumption into the loss function~\cite{yang2021circular}. However, because this method operates within Euclidean space and combines this term with other loss functions, it is highly probable that the resulting manifold structure does not converge to a true circle.

In this study, we design a loss function that induces a circular structure on a hypersphere, thereby explicitly reproducing this geometry within the language model's embedding space. This allows us to rigorously discuss the utility and validity of applying psychological circumplex models of emotion to deep learning.

\section{Methodology}
\subsection{Overview}
Figure~\ref{fig:Overview_framework} presents an overview of our experimental framework. We collect datasets where emotions are assumed to be equally spaced on a circle and train the model to encode their representations. To replicate the circular structure, we employ the following two learning strategies:

\noindent\textbf{Geometry of the embedding space.}
By learning emotion representations in a spherical space, we encourage the model to reproduce the circular structure, ensuring that differences are represented solely by angle distance (Section~\ref{ssec: architecture}).

\noindent\textbf{Contrastive learning design.}
We align the relationships between emotion labels by calibrating pairwise distances or gradient weights between anchor and negative samples according to their positions on the Circumplex Model of Affect (Section~\ref{ssec: loss function}).

\subsection{Preliminary}
\begin{figure}[tp]
    \includegraphics[width=1.0\linewidth]{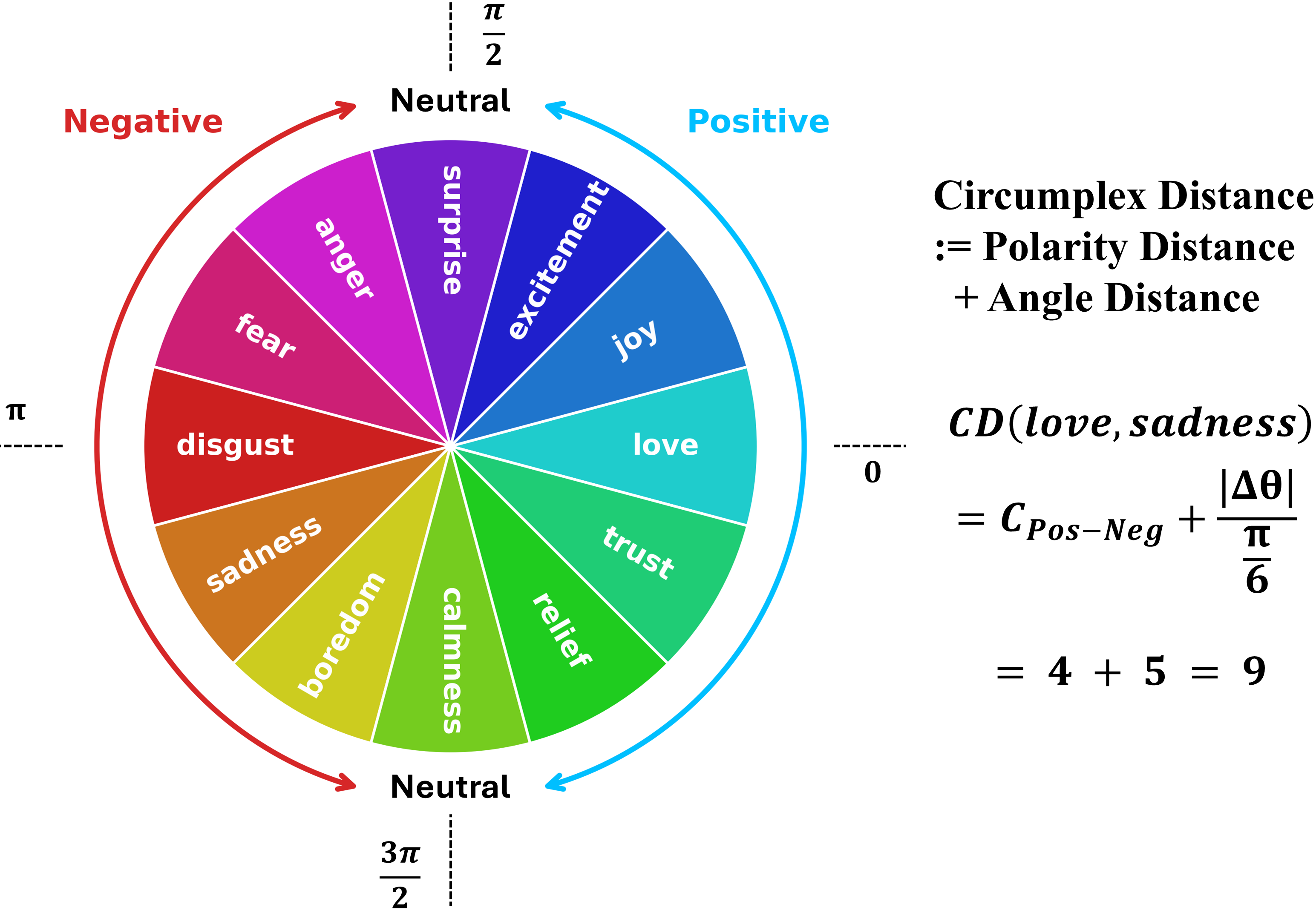}
    \caption{Our empirical circumplex model (ECM) and definition of Circumplex Distance (CD). }
    \label{fig:Emotion Wheel}
\end{figure}

Figure~\ref{fig:Emotion Wheel} illustrates the basic circumplex emotion arrangement employed in this study, which consists of 12 emotion categories (hereafter referred to as the Empirical Circumplex Model, or ECM). While our model is primarily based on Russell’s Circumplex Model of Affect~\cite{Russell1980, 12-point_circumplex}, certain emotions have been substituted to align with the labels available in real-world datasets. For instance, although terms like "sleepy" or "quiet" would ideally represent the state of deactivation at $3\pi/2$, we adopt "calmness" due to the absence of datasets annotated with these specific labels. Formally, we define each emotion label as $y \in \mathcal{E}$, where $y$ denotes the class of emotion and $\mathcal{E}$ denotes the complete set of emotion labels of size $E := |\mathcal{E}|$. In our experiments, we utilize a set of $N$ text-label pairs for a given emotion classification dataset, denoted as $D = \{(x_i, y_i) \mid y_i \in \mathcal{E}\}_{i=1}^N$. 
\subsection{Head Architecture}
\label{ssec: architecture}
We append a single Transformer block (projection head) to a pre-trained backbone, utilizing the head's output embedding space to represent the geometric structure of emotions. Let $t$ denote the index of the token in the input sequence $[1, \dots, T]$ and $d$ denote the dimension of the model. 
A standard Transformer block consists of an attention mechanism (ATTN), a multi-layer perceptron (MLP), and normalization modules (RMSNorm), formulated as follows:
\begin{equation}
    \begin{aligned}
    &h^{'}_t = h^{''}_t + \text{ATTN}(\text{RMSNorm}(h^{''}_t)), \\
    &h_t = h^{'}_t + \text{MLP}(\text{RMSNorm}(h^{'}_t)), \\
    \end{aligned}
\end{equation}
where $h_t, h^{'}_t, h^{''}_t \in \mathbb{R}^d$. Here, $h^{''}_t$ denotes the backbone output, which serves as the head input, and $h_t$ denotes the head output. This space corresponds to a $d$-dimensional Euclidean space where the arrangement of embeddings is unconstrained. Since the norm itself carries semantic meaning, it is inherently difficult to induce a circular geometry. Therefore, we adopt the normalized Transformer Block (nGPT), an architecture explicitly designed for spherical space~\cite{loshchilov2025ngpt}. The operations within the nGPT block are defined as follows:
\begin{equation} 
\scriptsize
\begin{aligned} 
&h^{'}_t = \text{Norm}((1-\alpha_A) \odot \text{Norm}(h_t^{''}) + \alpha_A \odot \text{Norm}(\text{ATTN}(h^{''}_t))), \\  
&h_t = \text{Norm}((1-\alpha_M) \odot \text{Norm}(h_t^{'}) + \alpha_M \odot \text{Norm}(\text{MLP}(h^{'}_t))), \\ 
\end{aligned} 
\end{equation}
where $\text{Norm()}$ represents $\ell_2$ normalization, $h_t, h^{'}_t \in \mathbb{S}^{d-1}$, and $\alpha_A, \alpha_M \in \mathbb{R}^{d}$ are learnable parameters. nGPT removes normalization modules from standard transformer blocks and instead normalizes all hidden states and weights to unit norm along the feature dimension. Consequently, since the output of each module resides on a hypersphere and is updated along geodesics, the optimization process within the block can be viewed as traversing the spherical manifold. The rationale for adopting the nGPT architecture and further details regarding the intra-block processing are provided in Appendices~\ref{sec: rationale for nGPT} and~\ref{sec: transformer block}. In our experiments, we derive the final sentence embedding $e$ by applying a pooling operation to the sequence of hidden states $h_{1:T}$, followed by normalization:
\begin{equation}
    e = \text{Norm}(\text{Pooling}(h_{1:T})),
\end{equation}
where $e \in \mathbb{S}^{d-1}, h_{1:T} \in \mathbb{R}^{T \times d}$.
The specific pooling operation depends on the backbone model and is defined as follows:
\begin{equation}
\small
\begin{aligned}    
\text{Pooling}_{\text{cls}} := h_1, \text{Pooling}_{\text{last}} := h_T, \text{Pooling}_{\text{mean}} := \frac{1}{T} \sum_{t=1}^{T} h_t.
\end{aligned}
\end{equation}
For comparison, a conventional Transformer block (GPT head) is also trained as a baseline. 
\subsection{Training Objective}
\label{ssec: loss function}
We perform contrastive learning by applying three distinct loss functions. Since our focus is on the quality of the manifold representation rather than emotion classification alone, we utilize the embeddings $e$ directly during evaluation. Consequently, we employ neither additional linear classification layers nor cross-entropy loss. This approach ensures that the distinct effects of each loss function are directly reflected in the characteristics of the embedding space. We employ the Supervised InfoNCE REvisited (SINCERE) loss~\cite{feeney2023sincere} as our baseline to achieve maximum discriminability while avoiding the problematic intra-class repulsion inherent in the original Supervised Contrastive Loss~\cite{khosla2020supervised}. It computes the mean loss across all positive pairs within the batch $\mathcal{B}$ of size $B := |\mathcal{B}|$ for a given anchor as follows:
\begin{equation}
\label{eq: sincere_overall}
\begin{aligned}
& \mathcal{L}_{\mathrm{SINCERE}} = \\
&\quad \frac{1}{B} \sum_{i \in \mathcal{B}}
\left(
\frac{-1}{|\mathcal{P}|}
\sum_{j \in \mathcal{P}}
\log
\frac{\exp(e^T_ie_j / \tau)}{Z_i}
\right)
\end{aligned}
\end{equation}
where $\tau$ is a temperature, $\mathcal{P}$ is the in-batch positive set of the anchor sentence $x_i$, and $Z_i$ represents the term corresponding to the positive sample as well as the in-batch negative samples:
\begin{equation}
\label{eq: sincere_negative_term}
Z_i = \exp(e^T_ie_j / \tau) + \sum_{k \in \mathcal{N}}\exp(e^T_ie_k / \tau)
\end{equation}
where $\mathcal{N}$ is the in-batch negative set. Since this loss function computes the loss for all pairs within the batch, it induces a strong separation between positive and negative samples. However, because it applies equal weight to all negative samples, it fails to account for the specific degree of separation required between the anchor and each individual negative sample. Given that the ECM positions similar emotions nearby and opposing emotions at antipodes, the loss function should ideally dictate specific pairwise distances. Accordingly, we utilize SoftCSE for soft constraints and CircularCSE for hard constraints. We refine the 'weight individualization' method proposed in previous SoftCSE research, as it offers a more intuitive formulation and greater numerical stability~\cite{zhuang2024not}. SoftCSE assigns individual weights to the negative sample terms in the SINCERE loss:
\begin{equation}
\small
Z_i = \exp(e^T_ie_j / \tau) + \sum_{k \in \mathcal{N}} w_{ik} \exp(e^T_ie_k / \tau),    
\end{equation}
\begin{equation}
    w_{ik} = \frac{1 - \cos({\Delta \theta_{i k}})}{\frac{1}{|\mathcal{N}|} \sum_{k \in \mathcal{N}} (1 - \cos(\Delta \theta_{i k}))}
\end{equation}
where $\Delta \theta_{ik}$ is the angular difference on the ECM. The numerator of $w_{ik}$ is inversely proportional to the similarity of the emotion labels $i$ and $k$ i.e., $w_{ik} \propto - \cos(\Delta \theta_{ik})$, meaning that the closer the labels are on the circle, the smaller the weight and the weaker the repelling force becomes. The denominator acts as a normalization factor within the batch, adjusting the scale of the negative terms to match Equation~(\ref{eq: sincere_negative_term}) i.e., $\sum_{k \in \mathcal{N}}1 = \sum_{k \in \mathcal{N}}w_{ik} = |\mathcal{N}|$. While the original paper employed $e^T_ie_k$ computed via a frozen encoder model instead of $\cos(\Delta \theta_{ik})$, this approach incurs additional inference overhead during training and relies heavily on the performance of the encoder. Therefore, we pre-define pairwise distances based on the ECM. This design facilitates the straightforward assignment of individual weights to negative samples. 

As an even stronger constraint, we propose CircularCSE, which directly learns distances on the circle:
\begin{equation}
\small
\begin{aligned}    
&\quad\quad\quad\mathcal{L}_{\text{CircularCSE}} = \frac{1}{B(B-1)} \sum_{i, j \in \mathcal{B}:i\neq j} \ell_{ij}, \\
&\ell_{ij} = 
\begin{cases} 
[\max(0, |e^T_ie_j  - \cos(\Delta \theta_{ij})| -m )]^2 & \text{if } y_i = y_j \\ 
(e^T_ie_j -\cos(\Delta \theta_{ij}))^2 & \text{otherwise}
\end{cases}
\end{aligned}
\end{equation}
Here, $m > 0$ is a margin hyperparameter that allows for tolerance within classes. Although a margin is desirable to account for variations in nuance and intensity among samples sharing the same label, it inevitably lowers the model's discriminability.

\begin{table*}[t]
\centering
\setlength{\tabcolsep}{11pt}
\begin{tabular}{lccccccc}
\toprule
\multirow{2}{*}{\shortstack[l]{Training\\Objective}} & \multirow{2}{*}{\shortstack[l]{Head\\Arch.}} & 
\multicolumn{2}{c}{mE5} & 
\multicolumn{2}{c}{\small Qwen3-Embedding-4B} & 
\multicolumn{2}{c}{Llama-3.2-3B} \\
\cmidrule(lr){3-4} \cmidrule(lr){5-6} \cmidrule(lr){7-8}
 & & $V_{\text{Measure}}$ & CD-r & $V_{\text{Measure}}$ & CD-r & $V_{\text{Measure}}$ & CD-r \\
\midrule
\multicolumn{2}{l}{Pretrained} & \multicolumn{1}{|c}{0.342} & 0.574 & \multicolumn{1}{|c}{0.495} & 0.522 & \multicolumn{1}{|c}{0.094} & 0.217 \\
\hline
\midrule
\multirow{2}{*}{\textbf{SINCERE}} 
 & - GPT & \multicolumn{1}{|c}{\textbf{0.760}} & 0.317 & \multicolumn{1}{|c}{\textbf{0.756}} & \uline{0.305} & \multicolumn{1}{|c}{\textbf{0.725}} & \uline{0.358} \\
 & - nGPT & \multicolumn{1}{|c}{0.744} & \uline{0.221} & \multicolumn{1}{|c}{0.739} & 0.545 & \multicolumn{1}{|c}{0.577} & 0.425 \\
\cmidrule(lr){1-2} 
\multirow{2}{*}{\textbf{SoftCSE}} 
 & - GPT & \multicolumn{1}{|c}{0.755} & 0.477 & \multicolumn{1}{|c}{0.751} & 0.552 & \multicolumn{1}{|c}{0.710} & 0.548 \\
 & - nGPT & \multicolumn{1}{|c}{0.753} & 0.499 & \multicolumn{1}{|c}{0.723} & 0.708 & \multicolumn{1}{|c}{0.516} & \textbf{0.728} \\
\cmidrule(lr){1-2} 
\multirow{2}{*}{\textbf{CircularCSE}} 
 & - GPT & \multicolumn{1}{|c}{\uline{0.717}} & 0.757 & \multicolumn{1}{|c}{\uline{0.643}} & 0.747 & \multicolumn{1}{|c}{0.579} & \textbf{0.728} \\
 & - nGPT & \multicolumn{1}{|c}{0.720} & \textbf{0.764} & \multicolumn{1}{|c}{0.659} & \textbf{0.753} & \multicolumn{1}{|c}{\uline{0.382}} & 0.708 \\
\bottomrule
\end{tabular}
\caption{Summary of average performance across all datasets for selected models. Best results per model (excluding Pretrained) are bolded, and worst are underlined. $V_{\text{Measure}}$ indicates clustering quality, and CD-r indicates correlation with the circumplex distance.}
\label{tab:performance_summary_selected}
\end{table*}
\section{Experiments}
\subsection{Experimental Setup}
\noindent\textbf{Datasets.} Our experiments utilize three real-world datasets: Emolit~\cite{app13137502}, Empathetic Dialogue~\cite{rashkin2019towards}, SuperEmotion~\cite{de2025super}, and one synthetic dataset: PersonaGen~\cite{inoshita2025persona}. Featuring a wide range of emotion labels, these datasets allow us to evaluate the geometric fidelity of the embedding space by analyzing how emotional categories are structurally organized. We select samples annotated with labels that match or closely resemble those in the ECM shown in Figure~\ref{fig:Emotion Wheel}. For each dataset, the training set consists of 500 instances sampled per emotion label (450 for Super-Emotion), and the test set comprises 100 instances per label. Detailed descriptions of each dataset and the construction pipeline for the synthetic dataset are provided in Appendix~\ref{sec: dataset collection}.

\noindent\textbf{Models.} To account for variations in embedding space properties resulting from diverse architectures and pre-training objectives, we categorize the models into three groups, selecting two representatives from each: \textbf{BERT-like Encoders} (mE5~\cite{wang2024multilingual} and mxbai~\cite{embed2024mxbai}), \textbf{LLM-based Encoders} (Qwen3-Embedding-4B~\cite{qwen3embedding} and Llama-Embed-Nemotron-8B~\cite{babakhin2025llamaembednemotron8buniversaltextembedding}), and \textbf{Decoder-only LLMs} (Llama-3.2-3B~\cite{grattafiori2024llama3herdmodels} and OLMo-3-7B~\cite{olmo3_2025}). 

\noindent\textbf{Implementation Details.} We attach a single-layer Transformer Block (GPT head) or a normalized Transformer Block (nGPT head) to the final layer of each backbone model. The head shares the same attention mask, dimension size, and attention head count as the backbone.
Final sentence embeddings are derived via specific pooling methods: the [CLS] token for BERT-like encoders, the last token for decoder-only LLMs and Qwen3-Embedding. For Llama-Embed-Nemotron, we utilize mean pooling, consistent with its original training methodology. During training, BERT-like encoders are fully fine-tuned, while the LLM backbones remain frozen with only the heads being trained. Further implementation details are provided in Appendix~\ref{sec: implementation details}.

\begin{figure*}[!pth]
  \centering
  \begin{subfigure}{0.24\linewidth}
    \includegraphics[width=\linewidth]{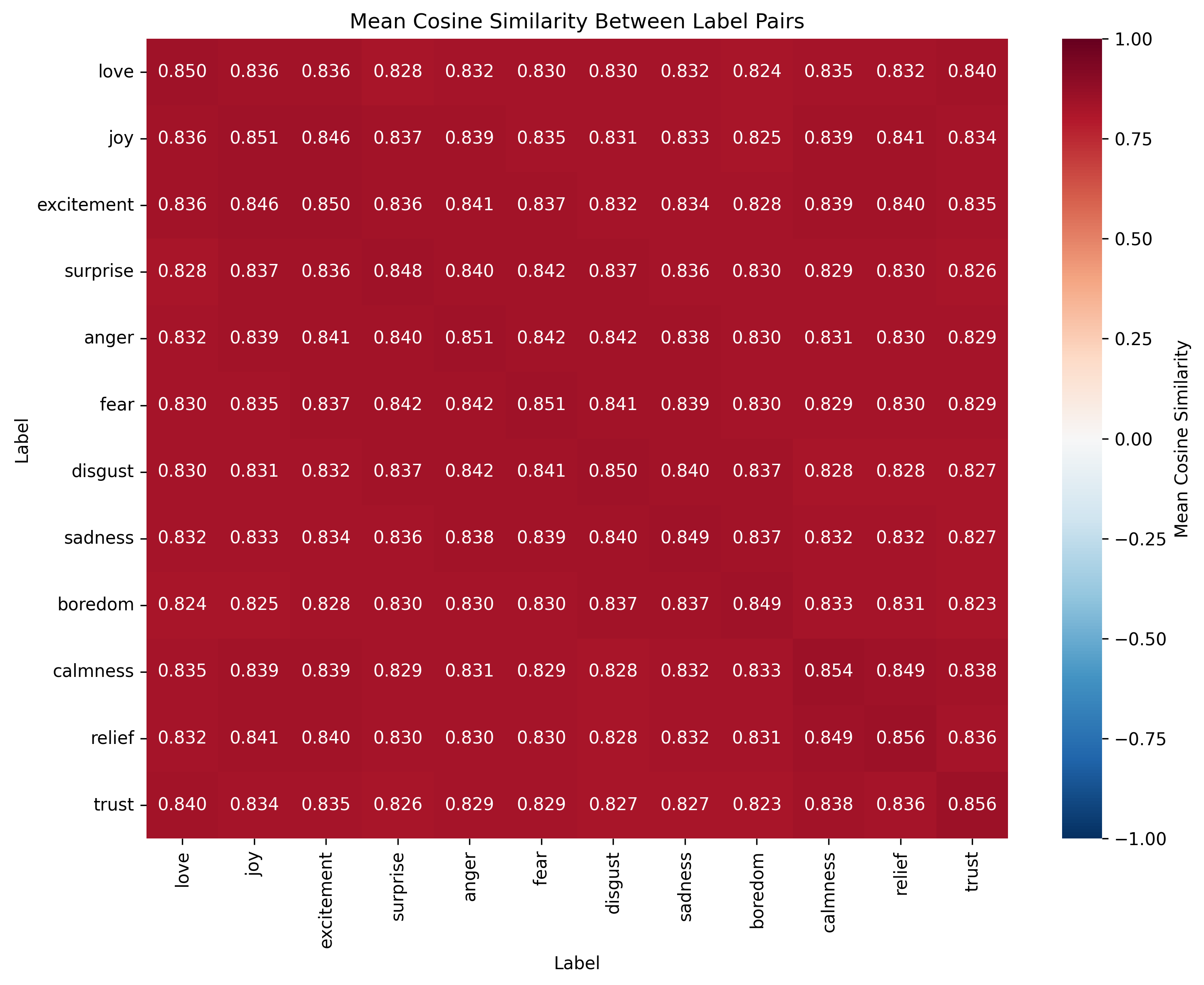}
    \caption{Pretrained}
  \end{subfigure}
  \hfill
  \begin{subfigure}{0.24\linewidth}
    \includegraphics[width=\linewidth]{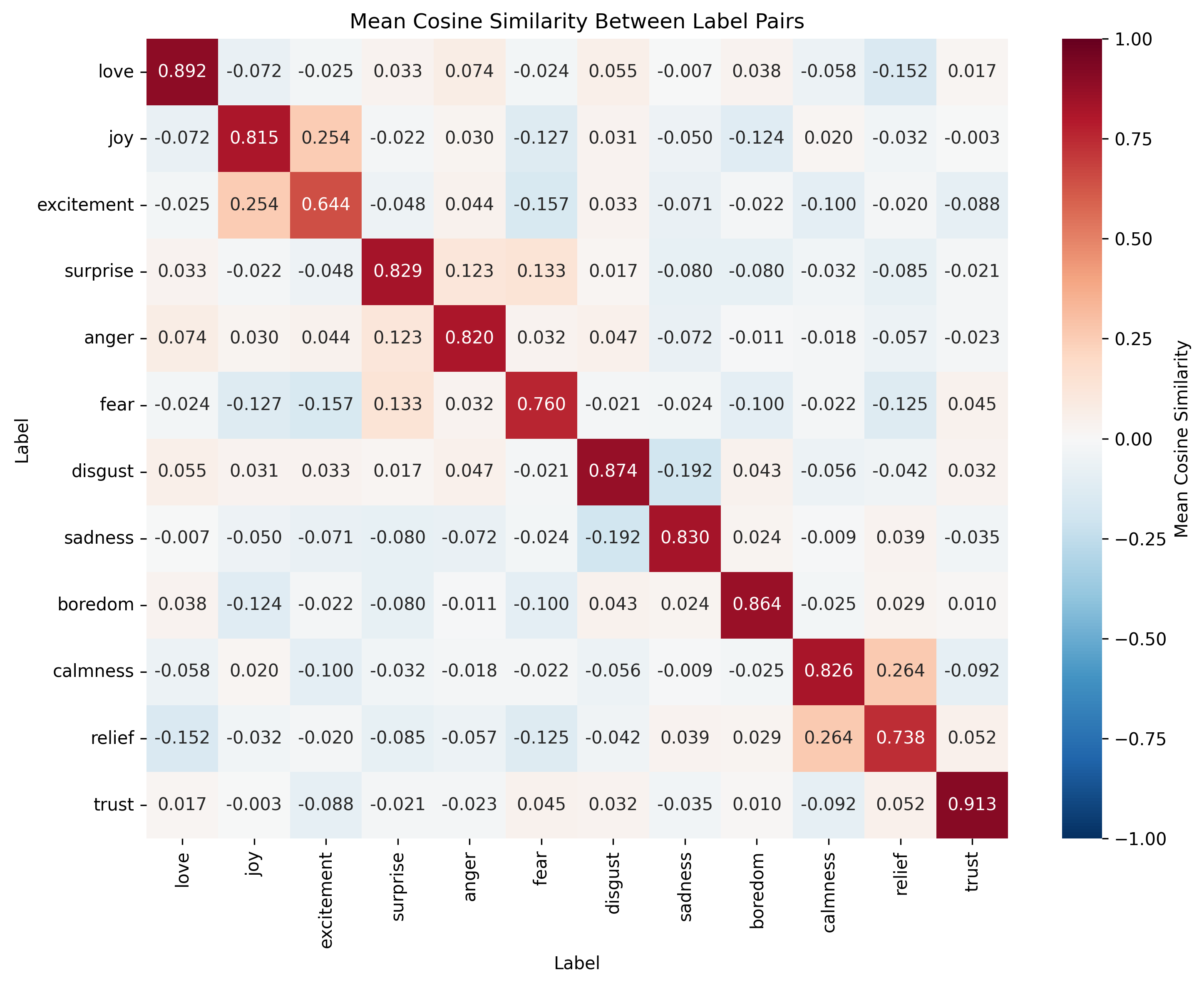}
    \caption{SINCERE-nGPT}
  \end{subfigure}
  \hfill
  \begin{subfigure}{0.24\linewidth}
    \includegraphics[width=\linewidth]{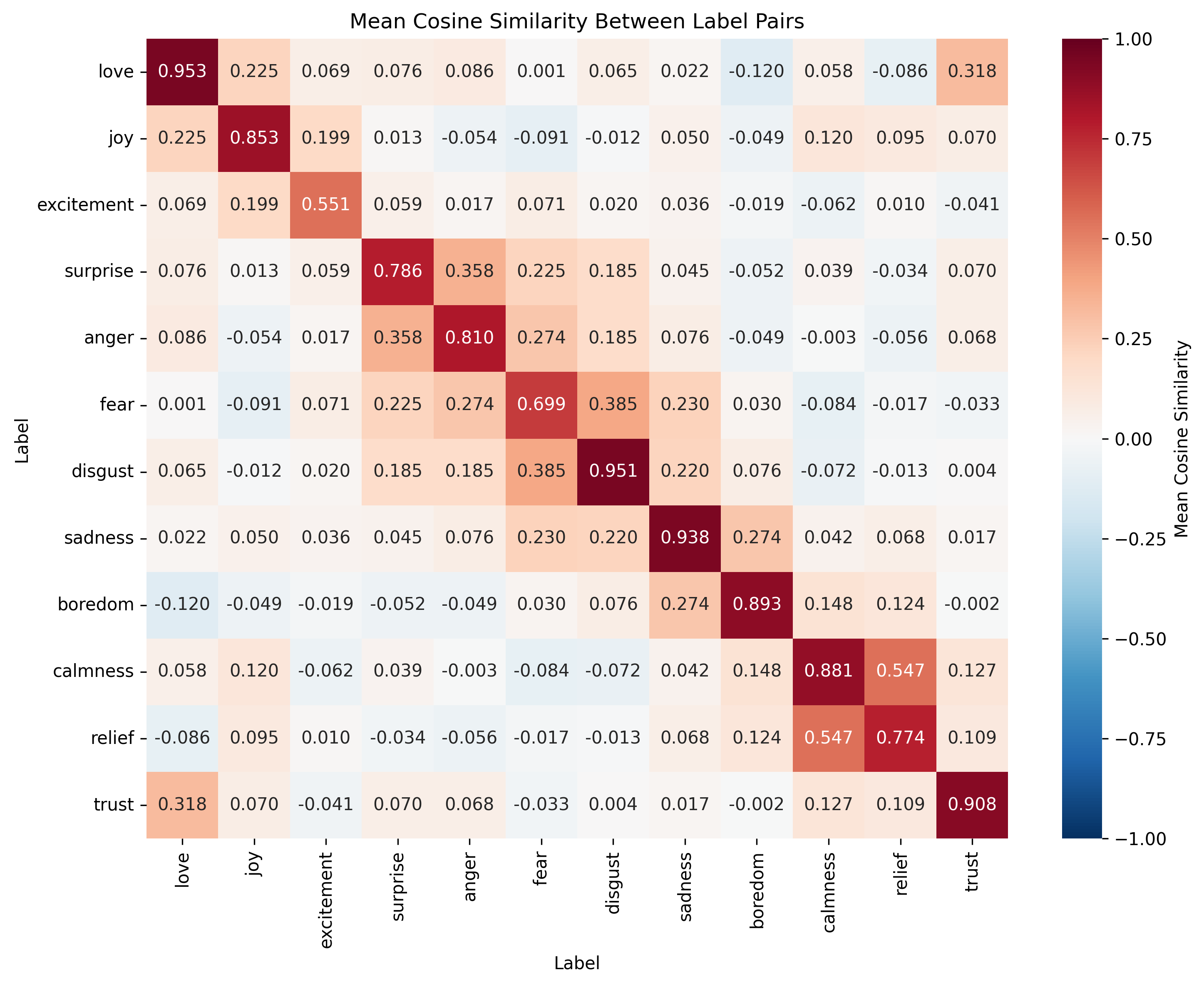}
    \caption{SoftCSE-nGPT}
  \end{subfigure}
  \hfill
  \begin{subfigure}{0.24\linewidth}
    \includegraphics[width=\linewidth]{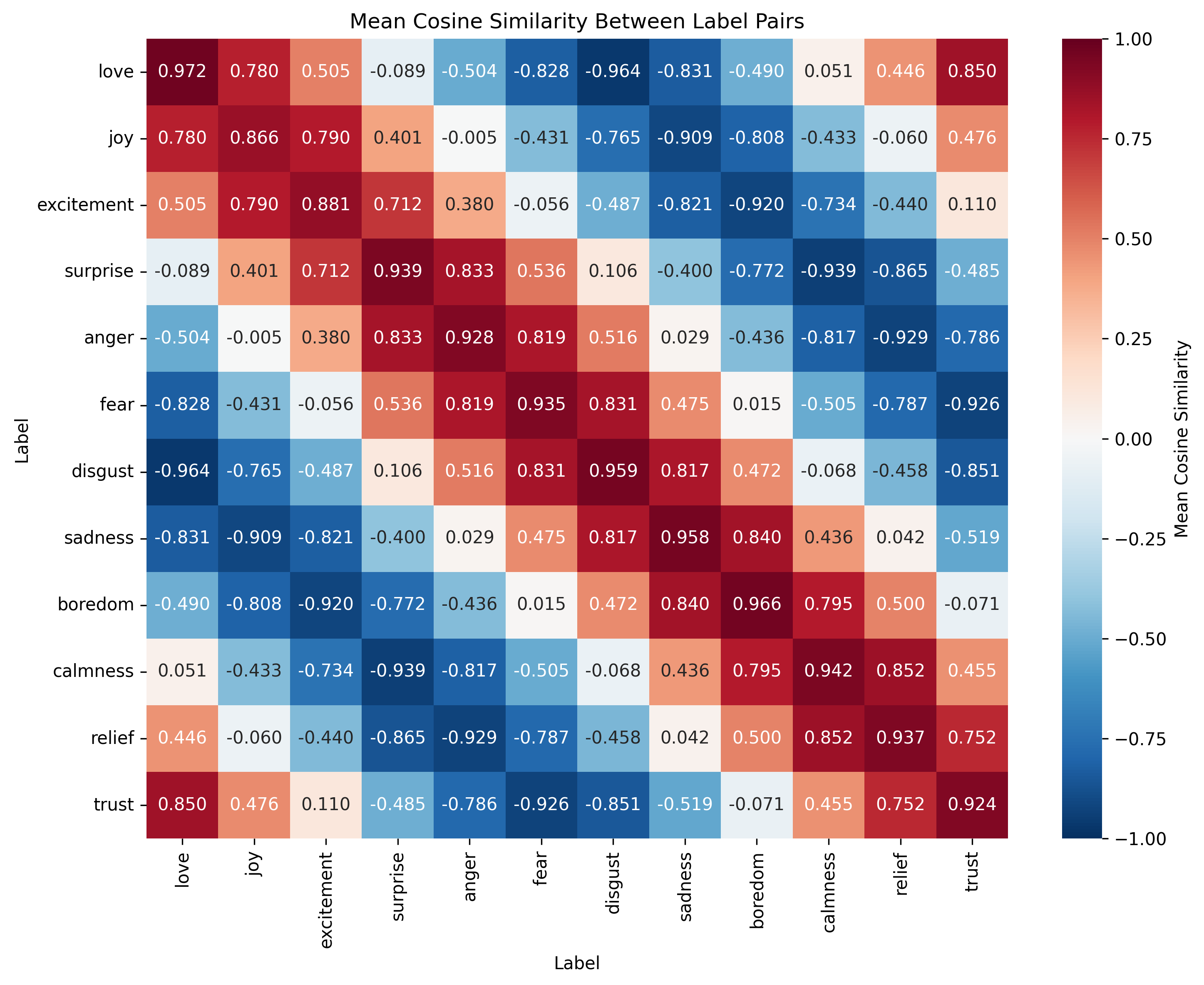}
    \caption{CircularCSE-nGPT}
  \end{subfigure}

  \caption{Average cosine similarity between emotion label pairs of mE5}
  \label{fig: AvgCosSim}
\end{figure*}

\subsection{Evaluation Metrics} We assess how well the trained embeddings capture emotional expressions via clustering analysis. Following the protocol of the MTEB clustering tasks~\cite{muennighoff-etal-2023-mteb}, we partition the test set into clusters using $k$-means~\cite{macqueen1967multivariate}, setting the number of clusters $k$ equal to the number of ground truth labels, and evaluate performance using V-Measure~\cite{rosenberg2007v}. Given that our models are optimized for cosine similarity, we utilize Spherical $k$-means~\cite{dhillon2001concept}, which uses cosine distance instead of the Euclidean $\ell_2$ metric. The algorithm is run 10 times with varied initializations, and the result with the minimum inertia is selected. We evaluate not only the discriminative power of the embedding space but also the extent to which its structure aligns with human perception. Building upon the approaches of~\cite{zhao2016predicting, zhao2024err}, we define the Circumplex Distance (CD) on the ECM as follows: $\text{CD}(y_i, y_j) := C + \text{AngleDistance}(y_i, y_j)$
where $C$ represents the constant inter-polarity distance, and $\text{AngleDistance}(y_i, y_j)$ denotes the number of steps between labels on the ECM. We define the distance between Neutral and Positive/Negative polarities as 2, the distance between Positive and Negative polarities as 4, and the distance between identical polarities as 0. This configuration ensures that the distance between opposing polarities consistently exceeds the distance between identical polarities. By incorporating the premise that polarity differences are cognitively more significant than mere steps on the circle, CD reflects human psychological emotion structure more faithfully.
To measure the alignment with this metric, we propose the Pearson correlation with CD (CD-r):
\begin{equation}
\small
    \text{CD-r} := \text{Pearson}( \text{CD}(y_i, y_j), 1 - \text{AvgCosSim}(y_i, y_j) )
\end{equation}
\begin{equation}
    \text{AvgCosSim}(y_i, y_j) := \frac{1}{N_i N_j}\sum_{k=1}^{N_i} \sum_{l=1}^{N_j} e^T_ke_l
\end{equation}
with $N_i$ and $N_j$ representing the total number of samples for each emotion label in the test set.

\subsection{Results}
\label{ssec: results}
Table~\ref{tab:performance_summary_selected} presents the average V-Measure and CD-r across datasets for mE5, Qwen3-Embedding-4B, and Llama-3.2-3B (comprehensive results are provided in Table~\ref{tab:performance_main_extended} of Appendix~\ref{sec: overall results}).

Regarding the training objectives, SINCERE and SoftCSE yield higher V-Measure scores, whereas CircularCSE underperforms. Conversely, for CD-r, SINCERE scores are low, while CircularCSE achieves higher results. This phenomenon can be intuitively explained by referencing Figure~\ref{fig: AvgCosSim}, which illustrates the average cosine similarity between emotion label pairs for each head.
Pre-trained models typically form an anisotropic embedding space, resulting in high similarity across all emotion label pairs~\cite{ethayarajh-2019-contextual}. In contrast, SINCERE attempts to position negative samples orthogonally, driving similarities toward zero. SoftCSE relaxes this constraint, and CircularCSE aligns the similarity of each emotion label pair with the corresponding cosine similarity on the circular manifold. Since clustering requires high discriminative power between labels, SINCERE is advantageous; however, it fails to sufficiently capture the relational structure between labels, leading to a lower CD-r. While CircularCSE successfully captures the ordinal relationships between labels, it makes distinguishing between adjacent labels more difficult, resulting in a lower V-Measure. These results highlight a fundamental conflict between the objectives of deep learning, which prioritizes discriminative accuracy, and psychology, which emphasizes alignment with human perception, creating a trade-off between accuracy and interpretability.

Revisiting Table~\ref{tab:performance_summary_selected}, we observe distinct trends across models; notably, for Llama-3.2-3B, the nGPT model exhibits a lower V-Measure. This suggests that Decoder-only models rely more heavily on vector norms to encode contextual information compared to Encoder models, implying that the normalization process leads to information loss. Furthermore, the significantly low accuracy of the pretrained models indicates that while emotion information is latent within the embeddings, it is not explicitly separable; instead, the embedding space prioritizes contextual information or features required for next-token prediction.
\begin{figure*}
    \centering
    \includegraphics[width=\linewidth]{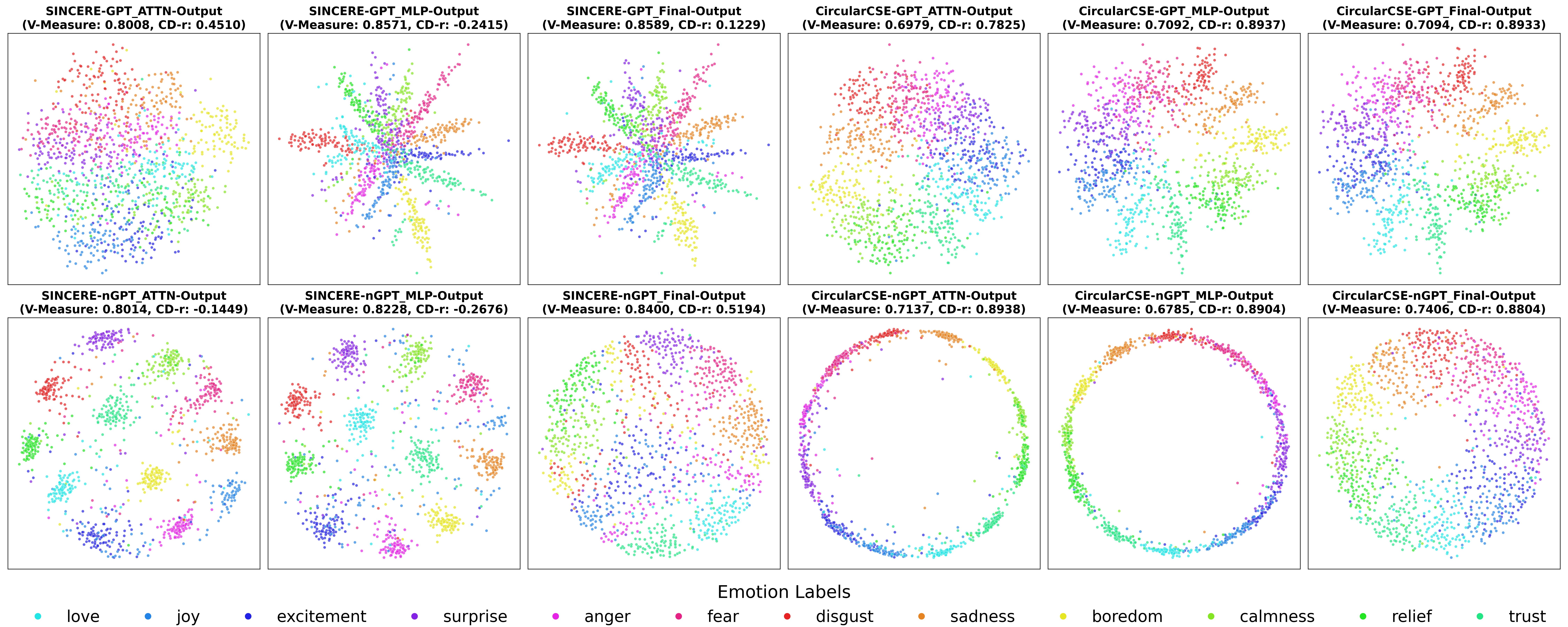}
    \caption{Visualization of embedding representations from each module of the GPT and nGPT heads using Multidimensional Scaling (MDS). Additional results for other models are presented in Appendix~\ref{sec: module visualization}.}
    \label{fig:main_mds_module}
\end{figure*}
\section{Analysis}
\subsection{Conflict between Psychological and Deep Learning Models}
To better elucidate the differences between the individual methods, we conduct further analysis. Figures~\ref{fig:pca2_visualization}(b), (c), and (d) visualize the results of applying PCA to the Emolit test set embeddings for each mE5 head. For SINCERE-nGPT and SoftCSE-nGPT, although the emotions separate into distinct clusters, the structures lack clear regularity, and the explained variance ratio of the principal components is low. In contrast, CircularCSE-nGPT exhibits a clear circular arrangement of emotions, with a significantly higher explained variance ratio. Hypothesizing that this difference in arrangement would manifest clearly under distinct scenarios, we conducted the following two experiments:

1. Robustness to dimensionality reduction.

2. Robustness to the number of emotion labels.
\begin{figure}[H] 
\centering
\includegraphics[width=\linewidth]{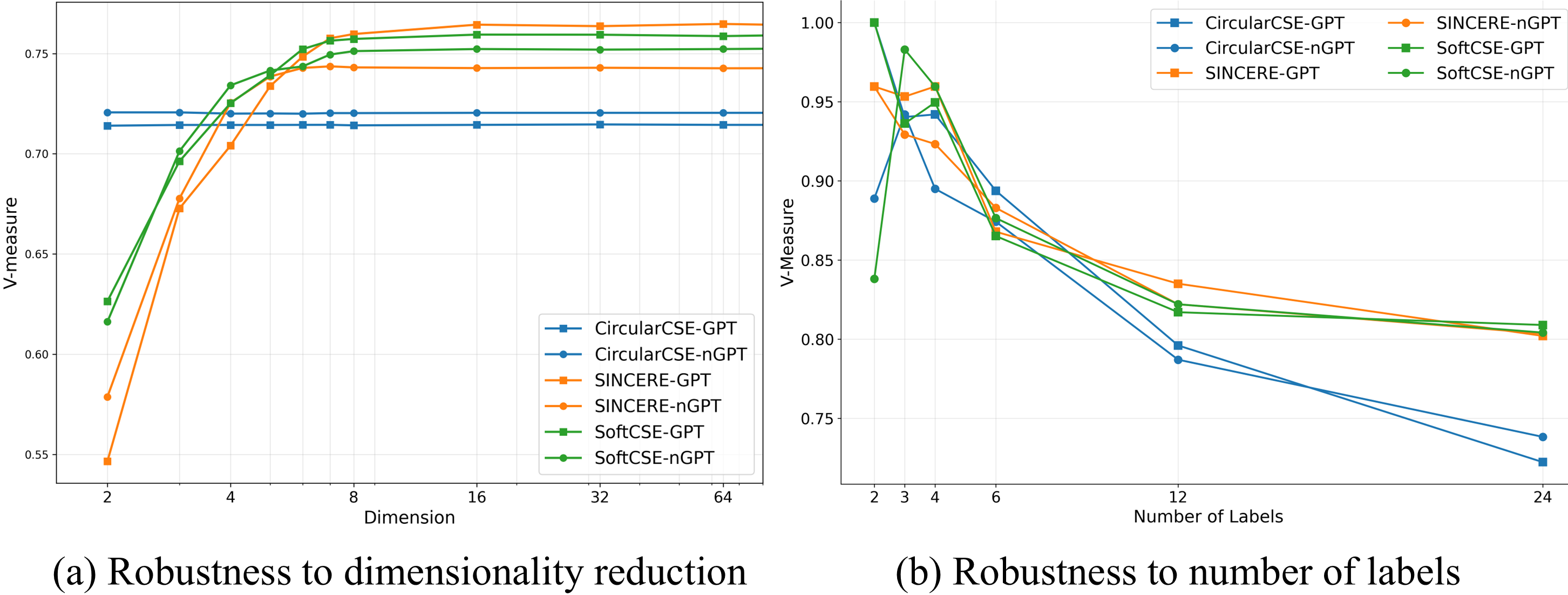}
\caption{Clustering performance of mE5 heads under different conditions. (a) Impact of PCA dimensionality reduction. (b) Impact of the number of emotion labels.}
\label{fig:robustness_experiment} 
\end{figure}
\noindent The experimental results are presented in Figure~\ref{fig:robustness_experiment}. Figure~\ref{fig:robustness_experiment}(a) shows the average V-Measure across datasets when clustering with Spherical $k$-means after reducing the dimensions of each mE5 head via PCA. Figure~\ref{fig:robustness_experiment}(b) illustrates the change in V-Measure on the Emolit dataset as the number of emotion labels varies (results for other models and details on label configurations are provided in Appendix~\ref{sec: robustness detail}). CircularCSE maintains stable accuracy even in low dimensions and performs comparably to SINCERE and SoftCSE when the number of emotion labels is small. However, its performance degrades in high-dimensional settings or with a large number of labels. These results can be explained by the optimal solutions and optimal margins of the respective loss functions. The SINCERE loss function, given by Equations~(\ref{eq: sincere_overall}) and (\ref{eq: sincere_negative_term}), attains its theoretical lower bound when the cosine similarity between all positive-negative pairs equals $\frac{-1}{E-1}$ (Proof in Appendix~\ref{sec: loss function math}). In practice, however, due to the curse of dimensionality, the positive-negative similarity often settles at a local optimum of 0, resulting in an average boundary margin of $90^\circ$ (orthogonality). In contrast, since CircularCSE arranges labels on a ring, the geometry of its optimal solution is consistently a 2-dimensional circle, and the boundary margin between any positive-negative pair is at most $\frac{\pi}{E}$ (for a 12-class classification problem, the maximum margin is $30^\circ$). The optimal boundary margins of SINCERE and CircularCSE coincide only in 2 dimensions; otherwise, the gap in discriminative power widens as the model dimensionality or the number of labels increases. This indicates that arranging emotions on a circular manifold, i.e., attempting to make representations visualization-friendly or semantically interpretable, implicitly imposes a low-dimensional manifold structure, which conflicts with the high-dimensional representations typical of deep learning. This limitation becomes more pronounced as model capacity increases (higher dimensionality) or as task demands rise (distinguishing among a diverse set of emotions).

\subsection{Effects of Spherical Constraints}
We perform a qualitative evaluation of the head architectural differences using dimensionality reduction. Specifically, we extract intermediate representations from within the transformer block of Qwen3-Embedding-4B and employ Multidimensional Scaling (MDS) for dimensionality reduction~\cite{articleMDS}, followed by clustering and visualization. Since MDS arranges points to reconstruct the pairwise distance matrix of the samples, it is particularly effective at preserving global geometry in high-dimensional spaces.

The results are presented in Figure~\ref{fig:main_mds_module}. The impact of the spherical constraint is prominently reflected in the structural differences of the representations. In the GPT head, the output of the MLP layer dominates the final result; the MLP amplifies differences in magnitude (norm), effectively overwriting prior intermediate representations with vectors of larger norms. Additionally, clusters exhibit linear elongation, where concepts appear as directional basis vectors. Although this expansion increases intra-cluster variance—allowing for a wider range of captured representations—it hinders Euclidean-based classification, as points may inadvertently lie closer to external clusters than to their own.

In contrast, in the nGPT head, the removal of the norm component results in the Attention and MLP layers exhibiting similar output geometries, with clusters adopting complex, non-linear structures rather than straight lines. We attribute this to the fact that on the hypersphere, differences between concepts are expressed solely via angular components rather than norms, causing directions to encode composite rather than singular concepts.

These findings suggest that even when quantitative evaluation metrics appear identical, the underlying representation structures can differ vastly. Manifold-aware research remains scarce, and determining the optimal manifold structure for specific tasks is an open question for future work.

\section{Discussion and Conclusion}
We compared conventional models with those reproducing psychological circular structures by performing contrastive learning in Euclidean and spherical spaces. Through this comparison, we discovered an unavoidable structural dilemma: a trade-off exists between discriminative performance and human interpretability. This highlights the importance of selecting the appropriate method based on the intended application. Although this study focused on clustering, our framework is transferable to representation learning for other concepts or tasks. For instance, interpretable approaches can potentially be applied to tasks such as identifying underrepresented labels and verifying the validity of emotion placement (Figure~\ref{fig:mds_all_label} in Appendix~\ref{sec: module visualization} provides an example).
Moreover, as interpretability and visualization facilitate model controllability, extending this approach to generation tasks is a key direction for future research.
\section*{Limitations}
\noindent\textbf{Task Simplification.} We simplified the task formulation for our experiments. Real-world emotions are highly complex and often involve simultaneous conflicting states (e.g., bittersweet, calm anger). However, in this study, we operate under the assumption that each text corresponds to a single emotion label. Furthermore, regarding the ``Neutral'' state, while we treated it as a distinct category belonging to neither Positive nor Negative extremes, it is intrinsically located at the center (origin) of Russell's circumplex model. Integrating these complex cases into our current framework would require specialized mechanisms; developing a more natural representation for such states remains a challenge for future work.

\noindent\textbf{Modality Constraints.} Emotion analysis spans not only text but also modalities such as images and audio. Certain emotional states (e.g., boredom or drowsiness) are more readily manifested through gestures or non-verbal cues rather than text. Capturing these nuances necessitates a multimodal architectural design.

\noindent\textbf{Connection to Psychological Models.} Although the circumplex model used in this study references Russell's model, some emotion placements differ from the original configuration to align with the specific labels available in our dataset. Additionally, numerous circumplex models exist beyond Russell's proposal, and it is possible that our specific arrangement does not perfectly reflect the true relationships between emotions (discussed in Appendix~\ref{ssec: label difference}). However, since our conclusions are largely driven by the dimensionality of the embedding space, we anticipate obtaining similar results with other 2- or 3-dimensional visualizable psychological models. Reproducing higher-dimensional and more complex emotional models is an objective for future research.

\section*{Ethical Considerations}
This study primarily utilizes publicly available and properly licensed datasets and models, all of which permit research use. The sole exception is the generation of synthetic data, where we employed both open-source and proprietary models to create a new dataset. While this dataset may potentially contain harmful content, it was used exclusively for training classification tasks; consequently, the trained models themselves pose no safety risks. All datasets and models were used in accordance with their intended research purposes.

We used AI tools for proofreading and mathematical typesetting in the preparation of this paper. All content has been thoroughly reviewed and verified by the human authors.
\section*{Acknowledgments}
This work was supported by Research and Development Center for Large Language Model/National Institute of Informatics, and JSPS KAKENHI Grant Number 24K03231.

\bibliography{latex/custom}

@inproceedings{yang2021circular,
  title={A circular-structured representation for visual emotion distribution learning},
  author={Yang, Jingyuan and Li, Jie and Li, Leida and Wang, Xiumei and Gao, Xinbo},
  booktitle={Proceedings of the IEEE/CVF Conference on Computer Vision and Pattern Recognition},
  pages={4237--4246},
  year={2021}
}

@article{zhao2024err,
  title={To err like human: Affective bias-inspired measures for visual emotion recognition evaluation},
  author={Zhao, Chenxi and Shi, Jinglei and Nie, Liqiang and Yang, Jufeng},
  journal={Advances in Neural Information Processing Systems},
  volume={37},
  pages={134747--134769},
  year={2024}
}

@article{12-point_circumplex,
  title={A 12-point circumplex structure of core affect.},
  author={Yik, Michelle and Russell, James A and Steiger, James H},
  journal={Emotion},
  volume={11},
  number={4},
  pages={705},
  year={2011},
  publisher={American Psychological Association}
}

@inproceedings{
loshchilov2025ngpt,
title={n{GPT}: Normalized Transformer with Representation Learning on the Hypersphere},
author={Ilya Loshchilov and Cheng-Ping Hsieh and Simeng Sun and Boris Ginsburg},
booktitle={The Thirteenth International Conference on Learning Representations},
year={2025},
url={https://openreview.net/forum?id=se4vjm7h4E}
}

@article{meng2019spherical,
  title={Spherical text embedding},
  author={Meng, Yu and Huang, Jiaxin and Wang, Guangyuan and Zhang, Chao and Zhuang, Honglei and Kaplan, Lance and Han, Jiawei},
  journal={Advances in neural information processing systems},
  volume={32},
  year={2019}
}

@misc{freenor2025steeringembeddingmodelsgeometric,
      title={Steering Embedding Models with Geometric Rotation: Mapping Semantic Relationships Across Languages and Models}, 
      author={Michael Freenor and Lauren Alvarez},
      year={2025},
      eprint={2510.09790},
      archivePrefix={arXiv},
      primaryClass={cs.CL},
      url={https://arxiv.org/abs/2510.09790}, 
}

@article{liu2022towards,
  title={Towards understanding grokking: An effective theory of representation learning},
  author={Liu, Ziming and Kitouni, Ouail and Nolte, Niklas S and Michaud, Eric and Tegmark, Max and Williams, Mike},
  journal={Advances in Neural Information Processing Systems},
  volume={35},
  pages={34651--34663},
  year={2022}
}

@inproceedings{
nanda2023progress,
title={Progress measures for grokking via mechanistic interpretability},
author={Neel Nanda and Lawrence Chan and Tom Lieberum and Jess Smith and Jacob Steinhardt},
booktitle={The Eleventh International Conference on Learning Representations },
year={2023},
url={https://openreview.net/forum?id=9XFSbDPmdW}
}

@article{khosla2020supervised,
  title={Supervised contrastive learning},
  author={Khosla, Prannay and Teterwak, Piotr and Wang, Chen and Sarna, Aaron and Tian, Yonglong and Isola, Phillip and Maschinot, Aaron and Liu, Ce and Krishnan, Dilip},
  journal={Advances in neural information processing systems},
  volume={33},
  pages={18661--18673},
  year={2020}
}

@Article{app13137502,
AUTHOR = {Rei, Luis and Mladenić, Dunja},
TITLE = {Detecting Fine-Grained Emotions in Literature},
JOURNAL = {Applied Sciences},
VOLUME = {13},
YEAR = {2023},
NUMBER = {13},
ARTICLE-NUMBER = {7502},
URL = {https://www.mdpi.com/2076-3417/13/13/7502},
ISSN = {2076-3417},
DOI = {10.3390/app13137502}
}

@article{de2025super,
  title={The Super Emotion Dataset},
  author={de Fortuny, Enric Junqu{\'e}},
  journal={arXiv preprint arXiv:2505.15348},
  year={2025}
}

@inproceedings{rashkin2019towards,
  title={Towards empathetic open-domain conversation models: A new benchmark and dataset},
  author={Rashkin, Hannah and Smith, Eric Michael and Li, Margaret and Boureau, Y-Lan},
  booktitle={Proceedings of the 57th annual meeting of the association for computational linguistics},
  pages={5370--5381},
  year={2019}
}

@article{inoshita2025persona,
  title={Persona-based synthetic data generation using multi-stage conditioning with large language models for emotion recognition},
  author={Inoshita, Keito and Harada, Rushia},
  journal={arXiv preprint arXiv:2507.13380},
  year={2025}
}

@article{wang2024multilingual,
  title={Multilingual E5 Text Embeddings: A Technical Report},
  author={Wang, Liang and Yang, Nan and Huang, Xiaolong and Yang, Linjun and Majumder, Rangan and Wei, Furu},
  journal={arXiv preprint arXiv:2402.05672},
  year={2024}
}

@online{embed2024mxbai,
  title={Open Source Strikes Bread - New Fluffy Embedding Model},
  author={Sean Lee and Aamir Shakir and Darius Koenig and Julius Lipp},
  year={2024},
  url={https://www.mixedbread.com/blog/mxbai-embed-large-v1},
}

@article{qwen3embedding,
  title={Qwen3 Embedding: Advancing Text Embedding and Reranking Through Foundation Models},
  author={Zhang, Yanzhao and Li, Mingxin and Long, Dingkun and Zhang, Xin and Lin, Huan and Yang, Baosong and Xie, Pengjun and Yang, An and Liu, Dayiheng and Lin, Junyang and Huang, Fei and Zhou, Jingren},
  journal={arXiv preprint arXiv:2506.05176},
  year={2025}
}

@misc{babakhin2025llamaembednemotron8buniversaltextembedding,
      title={Llama-Embed-Nemotron-8B: A Universal Text Embedding Model for Multilingual and Cross-Lingual Tasks}, 
      author={Yauhen Babakhin and Radek Osmulski and Ronay Ak and Gabriel Moreira and Mengyao Xu and Benedikt Schifferer and Bo Liu and Even Oldridge},
      year={2025},
      eprint={2511.07025},
      archivePrefix={arXiv},
      primaryClass={cs.CL},
      url={https://arxiv.org/abs/2511.07025}, 
}

@misc{grattafiori2024llama3herdmodels,
      title={The Llama 3 Herd of Models}, 
      author={Aaron Grattafiori and Abhimanyu Dubey and Abhinav Jauhri and Abhinav Pandey and Abhishek Kadian and Ahmad Al-Dahle and Aiesha Letman and Akhil Mathur and Alan Schelten and Alex Vaughan and Amy Yang and Angela Fan and Anirudh Goyal and Anthony Hartshorn and Aobo Yang and Archi Mitra and Archie Sravankumar and Artem Korenev and Arthur Hinsvark and Arun Rao and Aston Zhang and Aurelien Rodriguez and Austen Gregerson and Ava Spataru and Baptiste Roziere and Bethany Biron and Binh Tang and Bobbie Chern and Charlotte Caucheteux and Chaya Nayak and Chloe Bi and Chris Marra and Chris McConnell and Christian Keller and Christophe Touret and Chunyang Wu and Corinne Wong and Cristian Canton Ferrer and Cyrus Nikolaidis and Damien Allonsius and Daniel Song and Danielle Pintz and Danny Livshits and Danny Wyatt and David Esiobu and Dhruv Choudhary and Dhruv Mahajan and Diego Garcia-Olano and Diego Perino and Dieuwke Hupkes and Egor Lakomkin and Ehab AlBadawy and Elina Lobanova and Emily Dinan and Eric Michael Smith and Filip Radenovic and Francisco Guzmán and Frank Zhang and Gabriel Synnaeve and Gabrielle Lee and Georgia Lewis Anderson and Govind Thattai and Graeme Nail and Gregoire Mialon and Guan Pang and Guillem Cucurell and Hailey Nguyen and Hannah Korevaar and Hu Xu and Hugo Touvron and Iliyan Zarov and Imanol Arrieta Ibarra and Isabel Kloumann and Ishan Misra and Ivan Evtimov and Jack Zhang and Jade Copet and Jaewon Lee and Jan Geffert and Jana Vranes and Jason Park and Jay Mahadeokar and Jeet Shah and Jelmer van der Linde and Jennifer Billock and Jenny Hong and Jenya Lee and Jeremy Fu and Jianfeng Chi and Jianyu Huang and Jiawen Liu and Jie Wang and Jiecao Yu and Joanna Bitton and Joe Spisak and Jongsoo Park and Joseph Rocca and Joshua Johnstun and Joshua Saxe and Junteng Jia and Kalyan Vasuden Alwala and Karthik Prasad and Kartikeya Upasani and Kate Plawiak and Ke Li and Kenneth Heafield and Kevin Stone and Khalid El-Arini and Krithika Iyer and Kshitiz Malik and Kuenley Chiu and Kunal Bhalla and Kushal Lakhotia and Lauren Rantala-Yeary and Laurens van der Maaten and Lawrence Chen and Liang Tan and Liz Jenkins and Louis Martin and Lovish Madaan and Lubo Malo and Lukas Blecher and Lukas Landzaat and Luke de Oliveira and Madeline Muzzi and Mahesh Pasupuleti and Mannat Singh and Manohar Paluri and Marcin Kardas and Maria Tsimpoukelli and Mathew Oldham and Mathieu Rita and Maya Pavlova and Melanie Kambadur and Mike Lewis and Min Si and Mitesh Kumar Singh and Mona Hassan and Naman Goyal and Narjes Torabi and Nikolay Bashlykov and Nikolay Bogoychev and Niladri Chatterji and Ning Zhang and Olivier Duchenne and Onur Çelebi and Patrick Alrassy and Pengchuan Zhang and Pengwei Li and Petar Vasic and Peter Weng and Prajjwal Bhargava and Pratik Dubal and Praveen Krishnan and Punit Singh Koura and Puxin Xu and Qing He and Qingxiao Dong and Ragavan Srinivasan and Raj Ganapathy and Ramon Calderer and Ricardo Silveira Cabral and Robert Stojnic and Roberta Raileanu and Rohan Maheswari and Rohit Girdhar and Rohit Patel and Romain Sauvestre and Ronnie Polidoro and Roshan Sumbaly and Ross Taylor and Ruan Silva and Rui Hou and Rui Wang and Saghar Hosseini and Sahana Chennabasappa and Sanjay Singh and Sean Bell and Seohyun Sonia Kim and Sergey Edunov and Shaoliang Nie and Sharan Narang and Sharath Raparthy and Sheng Shen and Shengye Wan and Shruti Bhosale and Shun Zhang and Simon Vandenhende and Soumya Batra and Spencer Whitman and Sten Sootla and Stephane Collot and Suchin Gururangan and Sydney Borodinsky and Tamar Herman and Tara Fowler and Tarek Sheasha and Thomas Georgiou and Thomas Scialom and Tobias Speckbacher and Todor Mihaylov and Tong Xiao and Ujjwal Karn and Vedanuj Goswami and Vibhor Gupta and Vignesh Ramanathan and Viktor Kerkez and Vincent Gonguet and Virginie Do and Vish Vogeti and Vítor Albiero and Vladan Petrovic and Weiwei Chu and Wenhan Xiong and Wenyin Fu and Whitney Meers and Xavier Martinet and Xiaodong Wang and Xiaofang Wang and Xiaoqing Ellen Tan and Xide Xia and Xinfeng Xie and Xuchao Jia and Xuewei Wang and Yaelle Goldschlag and Yashesh Gaur and Yasmine Babaei and Yi Wen and Yiwen Song and Yuchen Zhang and Yue Li and Yuning Mao and Zacharie Delpierre Coudert and Zheng Yan and Zhengxing Chen and Zoe Papakipos and Aaditya Singh and Aayushi Srivastava and Abha Jain and Adam Kelsey and Adam Shajnfeld and Adithya Gangidi and Adolfo Victoria and Ahuva Goldstand and Ajay Menon and Ajay Sharma and Alex Boesenberg and Alexei Baevski and Allie Feinstein and Amanda Kallet and Amit Sangani and Amos Teo and Anam Yunus and Andrei Lupu and Andres Alvarado and Andrew Caples and Andrew Gu and Andrew Ho and Andrew Poulton and Andrew Ryan and Ankit Ramchandani and Annie Dong and Annie Franco and Anuj Goyal and Aparajita Saraf and Arkabandhu Chowdhury and Ashley Gabriel and Ashwin Bharambe and Assaf Eisenman and Azadeh Yazdan and Beau James and Ben Maurer and Benjamin Leonhardi and Bernie Huang and Beth Loyd and Beto De Paola and Bhargavi Paranjape and Bing Liu and Bo Wu and Boyu Ni and Braden Hancock and Bram Wasti and Brandon Spence and Brani Stojkovic and Brian Gamido and Britt Montalvo and Carl Parker and Carly Burton and Catalina Mejia and Ce Liu and Changhan Wang and Changkyu Kim and Chao Zhou and Chester Hu and Ching-Hsiang Chu and Chris Cai and Chris Tindal and Christoph Feichtenhofer and Cynthia Gao and Damon Civin and Dana Beaty and Daniel Kreymer and Daniel Li and David Adkins and David Xu and Davide Testuggine and Delia David and Devi Parikh and Diana Liskovich and Didem Foss and Dingkang Wang and Duc Le and Dustin Holland and Edward Dowling and Eissa Jamil and Elaine Montgomery and Eleonora Presani and Emily Hahn and Emily Wood and Eric-Tuan Le and Erik Brinkman and Esteban Arcaute and Evan Dunbar and Evan Smothers and Fei Sun and Felix Kreuk and Feng Tian and Filippos Kokkinos and Firat Ozgenel and Francesco Caggioni and Frank Kanayet and Frank Seide and Gabriela Medina Florez and Gabriella Schwarz and Gada Badeer and Georgia Swee and Gil Halpern and Grant Herman and Grigory Sizov and Guangyi and Zhang and Guna Lakshminarayanan and Hakan Inan and Hamid Shojanazeri and Han Zou and Hannah Wang and Hanwen Zha and Haroun Habeeb and Harrison Rudolph and Helen Suk and Henry Aspegren and Hunter Goldman and Hongyuan Zhan and Ibrahim Damlaj and Igor Molybog and Igor Tufanov and Ilias Leontiadis and Irina-Elena Veliche and Itai Gat and Jake Weissman and James Geboski and James Kohli and Janice Lam and Japhet Asher and Jean-Baptiste Gaya and Jeff Marcus and Jeff Tang and Jennifer Chan and Jenny Zhen and Jeremy Reizenstein and Jeremy Teboul and Jessica Zhong and Jian Jin and Jingyi Yang and Joe Cummings and Jon Carvill and Jon Shepard and Jonathan McPhie and Jonathan Torres and Josh Ginsburg and Junjie Wang and Kai Wu and Kam Hou U and Karan Saxena and Kartikay Khandelwal and Katayoun Zand and Kathy Matosich and Kaushik Veeraraghavan and Kelly Michelena and Keqian Li and Kiran Jagadeesh and Kun Huang and Kunal Chawla and Kyle Huang and Lailin Chen and Lakshya Garg and Lavender A and Leandro Silva and Lee Bell and Lei Zhang and Liangpeng Guo and Licheng Yu and Liron Moshkovich and Luca Wehrstedt and Madian Khabsa and Manav Avalani and Manish Bhatt and Martynas Mankus and Matan Hasson and Matthew Lennie and Matthias Reso and Maxim Groshev and Maxim Naumov and Maya Lathi and Meghan Keneally and Miao Liu and Michael L. Seltzer and Michal Valko and Michelle Restrepo and Mihir Patel and Mik Vyatskov and Mikayel Samvelyan and Mike Clark and Mike Macey and Mike Wang and Miquel Jubert Hermoso and Mo Metanat and Mohammad Rastegari and Munish Bansal and Nandhini Santhanam and Natascha Parks and Natasha White and Navyata Bawa and Nayan Singhal and Nick Egebo and Nicolas Usunier and Nikhil Mehta and Nikolay Pavlovich Laptev and Ning Dong and Norman Cheng and Oleg Chernoguz and Olivia Hart and Omkar Salpekar and Ozlem Kalinli and Parkin Kent and Parth Parekh and Paul Saab and Pavan Balaji and Pedro Rittner and Philip Bontrager and Pierre Roux and Piotr Dollar and Polina Zvyagina and Prashant Ratanchandani and Pritish Yuvraj and Qian Liang and Rachad Alao and Rachel Rodriguez and Rafi Ayub and Raghotham Murthy and Raghu Nayani and Rahul Mitra and Rangaprabhu Parthasarathy and Raymond Li and Rebekkah Hogan and Robin Battey and Rocky Wang and Russ Howes and Ruty Rinott and Sachin Mehta and Sachin Siby and Sai Jayesh Bondu and Samyak Datta and Sara Chugh and Sara Hunt and Sargun Dhillon and Sasha Sidorov and Satadru Pan and Saurabh Mahajan and Saurabh Verma and Seiji Yamamoto and Sharadh Ramaswamy and Shaun Lindsay and Shaun Lindsay and Sheng Feng and Shenghao Lin and Shengxin Cindy Zha and Shishir Patil and Shiva Shankar and Shuqiang Zhang and Shuqiang Zhang and Sinong Wang and Sneha Agarwal and Soji Sajuyigbe and Soumith Chintala and Stephanie Max and Stephen Chen and Steve Kehoe and Steve Satterfield and Sudarshan Govindaprasad and Sumit Gupta and Summer Deng and Sungmin Cho and Sunny Virk and Suraj Subramanian and Sy Choudhury and Sydney Goldman and Tal Remez and Tamar Glaser and Tamara Best and Thilo Koehler and Thomas Robinson and Tianhe Li and Tianjun Zhang and Tim Matthews and Timothy Chou and Tzook Shaked and Varun Vontimitta and Victoria Ajayi and Victoria Montanez and Vijai Mohan and Vinay Satish Kumar and Vishal Mangla and Vlad Ionescu and Vlad Poenaru and Vlad Tiberiu Mihailescu and Vladimir Ivanov and Wei Li and Wenchen Wang and Wenwen Jiang and Wes Bouaziz and Will Constable and Xiaocheng Tang and Xiaojian Wu and Xiaolan Wang and Xilun Wu and Xinbo Gao and Yaniv Kleinman and Yanjun Chen and Ye Hu and Ye Jia and Ye Qi and Yenda Li and Yilin Zhang and Ying Zhang and Yossi Adi and Youngjin Nam and Yu and Wang and Yu Zhao and Yuchen Hao and Yundi Qian and Yunlu Li and Yuzi He and Zach Rait and Zachary DeVito and Zef Rosnbrick and Zhaoduo Wen and Zhenyu Yang and Zhiwei Zhao and Zhiyu Ma},
      year={2024},
      eprint={2407.21783},
      archivePrefix={arXiv},
      primaryClass={cs.AI},
      url={https://arxiv.org/abs/2407.21783}, 
}

@techreport{olmo3_2025,
  title       = {OLMo 3 Technical Report},
  author      = {{OLMo Team} and Ettinger, Allyson and Heineman, David and Bertsch, Amanda and Kuehl, Bailey and Groeneveld, Dirk and Ivison, Hamish and others},
  institution = {Allen Institute for AI},
  year        = {2025},
  url         = {https://www.datocms-assets.com/64837/1763662397-1763646865-olmo_3_technical_report-1.pdf},
  note        = {Accessed: 2025-12-17}
}

@inproceedings{muennighoff-etal-2023-mteb,
    title = "{MTEB}: Massive Text Embedding Benchmark",
    author = "Muennighoff, Niklas  and
      Tazi, Nouamane  and
      Magne, Loic  and
      Reimers, Nils",
    editor = "Vlachos, Andreas  and
      Augenstein, Isabelle",
    booktitle = "Proceedings of the 17th Conference of the European Chapter of the Association for Computational Linguistics",
    month = may,
    year = "2023",
    address = "Dubrovnik, Croatia",
    publisher = "Association for Computational Linguistics",
    url = "https://aclanthology.org/2023.eacl-main.148/",
    doi = "10.18653/v1/2023.eacl-main.148",
    pages = "2014--2037"
}

@inproceedings{rosenberg2007v,
  title={V-measure: A conditional entropy-based external cluster evaluation measure},
  author={Rosenberg, Andrew and Hirschberg, Julia},
  booktitle={Proceedings of the 2007 joint conference on empirical methods in natural language processing and computational natural language learning (EMNLP-CoNLL)},
  pages={410--420},
  year={2007}
}

@inproceedings{macqueen1967multivariate,
  title={Multivariate observations},
  author={MacQueen, J},
  booktitle={Proceedings ofthe 5th Berkeley Symposium on Mathematical Statisticsand Probability},
  volume={1},
  pages={281--297},
  year={1967}
}

@article{dhillon2001concept,
  title={Concept decompositions for large sparse text data using clustering},
  author={Dhillon, Inderjit S and Modha, Dharmendra S},
  journal={Machine learning},
  volume={42},
  number={1},
  pages={143--175},
  year={2001},
  publisher={Springer}
}

@inproceedings{zhao2016predicting,
  title={Predicting personalized emotion perceptions of social images},
  author={Zhao, Sicheng and Yao, Hongxun and Gao, Yue and Ji, Rongrong and Xie, Wenlong and Jiang, Xiaolei and Chua, Tat-Seng},
  booktitle={Proceedings of the 24th ACM international conference on Multimedia},
  pages={1385--1394},
  year={2016}
}

@inproceedings{zhuang2024not,
  title={Not all negatives are equally negative: Soft contrastive learning for unsupervised sentence representations},
  author={Zhuang, Haojie and Emma Zhang, Wei and Yang, Jian and Chen, Weitong and Sheng, Quan Z},
  booktitle={Proceedings of the 33rd ACM International Conference on Information and Knowledge Management},
  pages={3591--3601},
  year={2024}
}

@article{zhao2024explainability,
  title={Explainability for large language models: A survey},
  author={Zhao, Haiyan and Chen, Hanjie and Yang, Fan and Liu, Ninghao and Deng, Huiqi and Cai, Hengyi and Wang, Shuaiqiang and Yin, Dawei and Du, Mengnan},
  journal={ACM Transactions on Intelligent Systems and Technology},
  volume={15},
  number={2},
  pages={1--38},
  year={2024},
  publisher={ACM New York, NY}
}

@article{
bereska2024mechanistic,
title={Mechanistic Interpretability for {AI} Safety - A Review},
author={Leonard Bereska and Stratis Gavves},
journal={Transactions on Machine Learning Research},
issn={2835-8856},
year={2024},
url={https://openreview.net/forum?id=ePUVetPKu6},
note={Survey Certification, Expert Certification}
}

@inproceedings{opitz-etal-2025-interpretable,
    title = "Interpretable Text Embeddings and Text Similarity Explanation: A Survey",
    author = "Opitz, Juri  and
      Moeller, Lucas  and
      Michail, Andrianos  and
      Pad{\'o}, Sebastian  and
      Clematide, Simon",
    editor = "Christodoulopoulos, Christos  and
      Chakraborty, Tanmoy  and
      Rose, Carolyn  and
      Peng, Violet",
    booktitle = "Proceedings of the 2025 Conference on Empirical Methods in Natural Language Processing",
    month = nov,
    year = "2025",
    address = "Suzhou, China",
    publisher = "Association for Computational Linguistics",
    url = "https://aclanthology.org/2025.emnlp-main.1135/",
    doi = "10.18653/v1/2025.emnlp-main.1135",
    pages = "22314--22330",
    ISBN = "979-8-89176-332-6"
}

@InProceedings{pmlr-v235-park24c,
  title = 	 {The Linear Representation Hypothesis and the Geometry of Large Language Models},
  author =       {Park, Kiho and Choe, Yo Joong and Veitch, Victor},
  booktitle = 	 {Proceedings of the 41st International Conference on Machine Learning},
  pages = 	 {39643--39666},
  year = 	 {2024},
  editor = 	 {Salakhutdinov, Ruslan and Kolter, Zico and Heller, Katherine and Weller, Adrian and Oliver, Nuria and Scarlett, Jonathan and Berkenkamp, Felix},
  volume = 	 {235},
  series = 	 {Proceedings of Machine Learning Research},
  month = 	 {21--27 Jul},
  publisher =    {PMLR},
  url = 	 {https://proceedings.mlr.press/v235/park24c.html}
}

@misc{elhage2022toymodelssuperposition,
      title={Toy Models of Superposition}, 
      author={Nelson Elhage and Tristan Hume and Catherine Olsson and Nicholas Schiefer and Tom Henighan and Shauna Kravec and Zac Hatfield-Dodds and Robert Lasenby and Dawn Drain and Carol Chen and Roger Grosse and Sam McCandlish and Jared Kaplan and Dario Amodei and Martin Wattenberg and Christopher Olah},
      year={2022},
      eprint={2209.10652},
      archivePrefix={arXiv},
      primaryClass={cs.LG},
      url={https://arxiv.org/abs/2209.10652}, 
}

@article{li2023inference,
  title={Inference-time intervention: Eliciting truthful answers from a language model},
  author={Li, Kenneth and Patel, Oam and Vi{\'e}gas, Fernanda and Pfister, Hanspeter and Wattenberg, Martin},
  journal={Advances in Neural Information Processing Systems},
  volume={36},
  pages={41451--41530},
  year={2023}
}

@article{Arditi2024RefusalIL,
  title={Refusal in Language Models Is Mediated by a Single Direction},
  author={Andy Arditi and Oscar Obeso and Aaquib Syed and Daniel Paleka and Nina Rimsky and Wes Gurnee and Neel Nanda},
  journal={ArXiv},
  year={2024},
  volume={abs/2406.11717},
  url={https://api.semanticscholar.org/CorpusID:270560489}
}

@inproceedings{
engels2025not,
title={Not All Language Model Features Are One-Dimensionally Linear},
author={Joshua Engels and Eric J Michaud and Isaac Liao and Wes Gurnee and Max Tegmark},
booktitle={The Thirteenth International Conference on Learning Representations},
year={2025},
url={https://openreview.net/forum?id=d63a4AM4hb}
}

@inproceedings{
park2025iclr,
title={{ICLR}: In-Context Learning of Representations},
author={Core Francisco Park and Andrew Lee and Ekdeep Singh Lubana and Yongyi Yang and Maya Okawa and Kento Nishi and Martin Wattenberg and Hidenori Tanaka},
booktitle={The Thirteenth International Conference on Learning Representations},
year={2025},
url={https://openreview.net/forum?id=pXlmOmlHJZ}
}

@article{Russell1980,
  author = {Russell, James A.},
  doi = {10.1037/h0077714},
  journal = {Journal of Personality and Social Psychology},
  month = {12},
  number = 6,
  pages = {1161--1178},
  publisher = {American Psychological Association},
  title = {A circumplex model of affect},
  volume = 39,
  year = 1980
}

@book{Plutchik1980Emotion,
  title={Emotion: A Psychoevolutionary Synthesis},
  author={Plutchik, Robert},
  year={1980},
  publisher={Harper \& Row},
  address={New York, USA}
}

@misc{zhao2025emergencehierarchicalemotionorganization,
      title={Emergence of Hierarchical Emotion Organization in Large Language Models}, 
      author={Bo Zhao and Maya Okawa and Eric J. Bigelow and Rose Yu and Tomer Ullman and Ekdeep Singh Lubana and Hidenori Tanaka},
      year={2025},
      eprint={2507.10599},
      archivePrefix={arXiv},
      primaryClass={cs.CL},
      url={https://arxiv.org/abs/2507.10599}, 
}

@book{Mehrabian1974AnAT,
  title     = {An Approach to Environmental Psychology},
  author    = {Mehrabian, Albert and Russell, James A.},
  year      = {1974},
  publisher = {The MIT Press},
  address   = {Cambridge, MA},
  isbn      = {978-0262130905}
}

@article{shaver1987emotion,
  title={Emotion knowledge: further exploration of a prototype approach.},
  author={Shaver, Phillip and Schwartz, Judith and Kirson, Donald and O'connor, Cary},
  journal={Journal of personality and social psychology},
  volume={52},
  number={6},
  pages={1061},
  year={1987},
  publisher={American Psychological Association}
}

@article{ekman1992argument,
  title={An argument for basic emotions},
  author={Ekman, Paul},
  journal={Cognition \& emotion},
  volume={6},
  number={3-4},
  pages={169--200},
  year={1992},
  publisher={Taylor \& Francis}
}

@inproceedings{guo2021enhancing,
  title={Enhancing cognitive models of emotions with representation learning},
  author={Guo, Yuting and Choi, Jinho D},
  booktitle={Proceedings of the Workshop on Cognitive Modeling and Computational Linguistics},
  pages={141--148},
  year={2021}
}

@ARTICLE{10902477,
  author={Wang, Xiangyu and Zong, Chengqing},
  journal={IEEE Transactions on Pattern Analysis and Machine Intelligence}, 
  title={Learning Emotion Category Representation to Detect Emotion Relations Across Languages}, 
  year={2025},
  volume={47},
  number={6},
  pages={4752-4767},
  doi={10.1109/TPAMI.2025.3545447}}

@article{reichman2025emotions,
  title={Emotions Where Art Thou: Understanding and Characterizing the Emotional Latent Space of Large Language Models},
  author={Reichman, Benjamin and Avsian, Adar and Heck, Larry},
  journal={arXiv preprint arXiv:2510.22042},
  year={2025}
}

@misc{choi2020amclossangularmargincontrastive,
      title={AMC-Loss: Angular Margin Contrastive Loss for Improved Explainability in Image Classification}, 
      author={Hongjun Choi and Anirudh Som and Pavan Turaga},
      year={2020},
      eprint={2004.09805},
      archivePrefix={arXiv},
      primaryClass={cs.CV},
      url={https://arxiv.org/abs/2004.09805}, 
}

@article{Deng_2022,
   title={ArcFace: Additive Angular Margin Loss for Deep Face Recognition},
   volume={44},
   ISSN={1939-3539},
   url={http://dx.doi.org/10.1109/TPAMI.2021.3087709},
   DOI={10.1109/tpami.2021.3087709},
   number={10},
   journal={IEEE Transactions on Pattern Analysis and Machine Intelligence},
   publisher={Institute of Electrical and Electronics Engineers (IEEE)},
   author={Deng, Jiankang and Guo, Jia and Yang, Jing and Xue, Niannan and Kotsia, Irene and Zafeiriou, Stefanos},
   year={2022},
   month=oct, pages={5962–5979} }

@article{bricken2023monosemanticity,
   title={Towards Monosemanticity: Decomposing Language Models With Dictionary Learning},
   author={Bricken, Trenton and Templeton, Adly and Batson, Joshua and Chen, Brian and Jermyn, Adam and Conerly, Tom and Turner, Nick and Anil, Cem and Denison, Carson and Askell, Amanda and Lasenby, Robert and Wu, Yifan and Kravec, Shauna and Schiefer, Nicholas and Maxwell, Tim and Joseph, Nicholas and Hatfield-Dodds, Zac and Tamkin, Alex and Nguyen, Karina and McLean, Brayden and Burke, Josiah E and Hume, Tristan and Carter, Shan and Henighan, Tom and Olah, Christopher},
   year={2023},
   journal={Transformer Circuits Thread},
   note={https://transformer-circuits.pub/2023/monosemantic-features/index.html}
    }

@inproceedings{ethayarajh-2019-contextual,
    title = "How Contextual are Contextualized Word Representations? {C}omparing the Geometry of {BERT}, {ELM}o, and {GPT}-2 Embeddings",
    author = "Ethayarajh, Kawin",
    editor = "Inui, Kentaro  and
      Jiang, Jing  and
      Ng, Vincent  and
      Wan, Xiaojun",
    booktitle = "Proceedings of the 2019 Conference on Empirical Methods in Natural Language Processing and the 9th International Joint Conference on Natural Language Processing (EMNLP-IJCNLP)",
    month = nov,
    year = "2019",
    address = "Hong Kong, China",
    publisher = "Association for Computational Linguistics",
    url = "https://aclanthology.org/D19-1006/",
    doi = "10.18653/v1/D19-1006",
    pages = "55--65"
}

@article{feeney2023sincere,
  title={Sincere: Supervised information noise-contrastive estimation revisited},
  author={Feeney, Patrick and Hughes, Michael C},
  journal={arXiv preprint arXiv:2309.14277},
  year={2023}
}

@article{articleMDS,
author = {de Leeuw, Jan},
year = {2005},
month = {10},
pages = {},
title = {Modern Multidimensional Scaling: Theory and Applications (Second Edition)},
volume = {14},
journal = {Journal of Statistical Software},
doi = {10.18637/jss.v014.b04}
}

@article{demszky2020goemotions,
  title={GoEmotions: A dataset of fine-grained emotions},
  author={Demszky, Dorottya and Movshovitz-Attias, Dana and Ko, Jeongwoo and Cowen, Alan and Nemade, Gaurav and Ravi, Sujith},
  journal={arXiv preprint arXiv:2005.00547},
  year={2020}
}

@inproceedings{van2012designing,
  title={Designing a scalable crowdsourcing platform},
  author={Van Pelt, Chris and Sorokin, Alex},
  booktitle={Proceedings of the 2012 ACM SIGMOD International Conference on Management of Data},
  pages={765--766},
  year={2012}
}

@misc{gpt5nano,
  author       = {{OpenAI}},
  title        = {GPT-5 Nano Model Overview},
  year         = {2025},
  howpublished = {\url{https://openai.com/ja-JP/index/gpt-5-system-card/}},
  note         = {Accessed: 2025-12-27},
}

@misc{olmo3_7b_instruct,
  author       = {{Allen Institute for AI}},
  title        = {Olmo-3-7B-Instruct Model Card},
  year         = {2025},
  howpublished = {\url{https://huggingface.co/allenai/Olmo-3-7B-Instruct}},
  note         = {Accessed: 2025-12-27},
}

@misc{qwen3_30b_a3b_instruct_2507,
  author       = {{Qwen}},
  title        = {Qwen3-30B-A3B-Instruct-2507 Model Card},
  year         = {2025},
  howpublished = {\url{https://huggingface.co/Qwen/Qwen3-30B-A3B-Instruct-2507}},
  note         = {Accessed: 2025-12-27},
}

@misc{phi4_mini_instruct,
  author       = {{Microsoft}},
  title        = {Phi-4-mini-instruct Model Card},
  year         = {2025},
  howpublished = {\url{https://huggingface.co/microsoft/Phi-4-mini-instruct}},
  note         = {Accessed: 2025-12-27},
}

@misc{llama3_3_70b_instruct,
  author       = {{Meta AI}},
  title        = {Llama-3.3-70B-Instruct Model Card},
  year         = {2024},
  howpublished = {\url{https://www.llama.com/docs/model-cards-and-prompt-formats/llama3_3/}},
  note         = {Accessed: 2025-12-27},
}

@misc{google2025gemma3,
  author ={Google DeepMind},
  title ={gemma3-27b-it Model Card},
  year ={2025},
  howpublished =
{\url{https://huggingface.co/google/gemma-3-27b-it}},
note = {Accessed: 2025-12-17},
}

@misc{ministral3_14b,
  author       = {{Mistral AI}},
  title        = {Ministral-3-14B Model Card},
  year         = {2025},
  howpublished = {\url{https://huggingface.co/mistralai/Ministral-3-14B-Base-2512}},
  note         = {Accessed: 2025-12-27},
}

@article{
doi:10.1073/pnas.2015509117,
author = {Vardan Papyan  and X. Y. Han  and David L. Donoho },
title = {Prevalence of neural collapse during the terminal phase of deep learning training},
journal = {Proceedings of the National Academy of Sciences},
volume = {117},
number = {40},
pages = {24652-24663},
year = {2020},
doi = {10.1073/pnas.2015509117},
URL = {https://www.pnas.org/doi/abs/10.1073/pnas.2015509117},
eprint = {https://www.pnas.org/doi/pdf/10.1073/pnas.2015509117}}

@InProceedings{pmlr-v139-graf21a,
  title = 	 {Dissecting Supervised Contrastive Learning},
  author =       {Graf, Florian and Hofer, Christoph and Niethammer, Marc and Kwitt, Roland},
  booktitle = 	 {Proceedings of the 38th International Conference on Machine Learning},
  pages = 	 {3821--3830},
  year = 	 {2021},
  editor = 	 {Meila, Marina and Zhang, Tong},
  volume = 	 {139},
  series = 	 {Proceedings of Machine Learning Research},
  month = 	 {18--24 Jul},
  publisher =    {PMLR},
  url = 	 {https://proceedings.mlr.press/v139/graf21a.html}
}
\appendix

\section{Justification to use nGPT}
\label{sec: rationale for nGPT}
In this section, we provide a detailed rationale for adopting the nGPT architecture to reproduce the circular structure within the model's embedding space. Simply, normalizing the model's embeddings at every step would satisfy the condition; this ensures all outputs lie on the unit hypersphere, meaning differences are expressed solely as angular differences. However, this design introduces significant issues for gradient descent during training.

Since the hypersphere is a non-Euclidean space where distance is geodesic, the standard linear vector addition inherent in residual connections updates vectors in the direction of steepest descent without accounting for curvature. Consequently, the updated representations deviate from (or 'fall off') the hypersphere. Therefore, we must employ Riemannian optimization instead of standard gradient descent to account for the manifold's curvature during training~\cite{meng2019spherical, freenor2025steeringembeddingmodelsgeometric}, necessitating the design of a more sophisticated architecture. The nGPT architecture is particularly compatible with representation learning for the following three reasons:

\textbf{Elimination of Weight Decay:} Research on grokking suggests that weight decay is critical for the emergence of generalization circuits~\cite{liu2022towards, nanda2023progress}. Models often memorize solutions within the magnitude (norm) of the weights; thus, weight decay is typically required to suppress these norm components and encourage structural generalization. In nGPT, however, weights are normalized by design, compelling the network to learn generalized structures directly without relying on weight decay.

\textbf{Feasibility of Head Training:} To the best of our knowledge, pre-trained weights for full-scale nGPT models are not currently available, and training from scratch is computationally expensive. However, in embedding tasks, it is standard practice to freeze the backbone and train only a projection head. Adopting an nGPT head aligns with this convention while bypassing the cost of full pre-training.    

\textbf{Angle-based Optimization:} In the original paper, nGPT was evaluated primarily as a causal language model, focusing on convergence speed rather than the implications of learning on a spherical manifold. In embedding tasks, however, cosine similarity (i.e., distance on the unit hypersphere) is the standard metric. Therefore, nGPT enables consistent angle-based optimization of representations throughout both training and inference.

These advantages make this architecture highly effective, not only for the current experiment but also as a general design principle for training embedding models.

\section{Transformer Block}
\label{sec: transformer block}
This section describes the detailed computational flow of the GPT head and the nGPT head adopted in this paper.
Let $t$ denote the index of the token in the input sequence $[1, \dots, T]$ and $d$ denote the dimension of the model. 
A standard Transformer block consists of an attention mechanism (ATTN), a multi-layer perceptron (MLP), and normalization modules (RMSNorm), formulated as follows:
\begin{equation}
    \begin{aligned}
    &h^{'}_t = h^{''}_t + \text{ATTN}(\text{RMSNorm}(h^{''}_t)), \\
    &h_t = h^{'}_t + \text{MLP}(\text{RMSNorm}(h^{'}_t)), \\
    \end{aligned}
\end{equation}
where $h_t, h^{'}_t, h^{''}_t \in \mathbb{R}^d$. Here, $h^{''}_t$ denotes the input to the block, and $h_t$ denotes the output. In our experiments, we derive the final sentence embedding $e$ by applying a pooling operation to the sequence of hidden states $h_{1:T}$, followed by normalization:
\begin{equation}
    e = \text{Norm}(\text{Pooling}(h_{1:T})),
\end{equation}
where $e \in \mathbb{S}^{d-1}, h_{1:T} \in \mathbb{R}^{T \times d}$.
The specific pooling operation depends on the backbone model and is defined as follows:
\begin{equation}
\begin{aligned}    
&\text{Pooling}_{\text{cls}} := h_1, \text{Pooling}_{\text{last}} := h_T,  \\
&\text{Pooling}_{\text{mean}} := \frac{1}{T} \sum_{t=1}^{T} h_t, 
\end{aligned}
\end{equation}
The ATTN performs the following computations:
\begin{equation}
    \text{ATTN}(h^{''}_t; \ h^{''}_{1:t}) = \text{Concat}(\text{A}_1,...,\text{A}_{n_{heads}})\boldsymbol{W}_O,\\
\end{equation}
\begin{equation}
\small    
    \text{A}_n =  \text{softmax}\left(\frac{(h^{''}_t\boldsymbol{W}_Q^n)(h^{''}_{1:t}\boldsymbol{W}_K^n)^T}{\sqrt{d_{head}}}\right) \ (h^{''}_{1:t}\boldsymbol{W}_V^n),
\end{equation}
where $\boldsymbol{W}_Q^n, \boldsymbol{W}_K^n , \boldsymbol{W}_V^n \in \mathbb{R}^{d \times d_{head}}, \boldsymbol{W}_O \in \mathbb{R}^{d \times d}, d_{head} = d / n_{heads}.$ For unidirectional models, the attention mask references tokens from $1$ to $t$, whereas for bidirectional models, it references the entire sequence from $1$ to $T$. The MLP operation is defined as follows
\begin{equation}
\text{MLP}(h^{'}_t) = (\text{SiLU}(h^{'}_t\boldsymbol{W}_u) \odot h^{'}_t\boldsymbol{W}_v)\boldsymbol{W}_{o\text{MLP}},
\end{equation}
where $\boldsymbol{W}_u, \boldsymbol{W}_v \in \mathbb{R}^{d \times d_{\text{MLP}}}, \boldsymbol{W_{\text{oMLP}}} \in \mathbb{R}^{d_{\text{MLP}} \times d}.$
Next, we describe the architecture of the Normalized Transformer Block (nGPT). The first primary modification is that residual connections are computed along geodesics:
\begin{equation} 
\scriptsize
\begin{aligned} 
&h^{'}_t = \text{Norm}((1-\alpha_A) \odot \text{Norm}(h_t^{''}) + \alpha_A \odot \text{Norm}(\text{ATTN}(h^{''}_t))), \\ 
&h_t = \text{Norm}((1-\alpha_M) \odot \text{Norm}(h_t^{'}) + \alpha_M \odot \text{Norm}(\text{MLP}(h^{'}_t))), \\ 
\end{aligned} 
\end{equation}
where $h_t, h^{'}_t, h^{''}_t \in \mathbb{S}^{d-1}$, and $\alpha_A, \alpha_M \in \mathbb{R}^{d}$ are learnable parameters. By constraining the vector updates to the hypersphere, this process enables pseudo-Riemannian optimization.
The ATTN and MLP modules incorporate the following modifications:
\begin{equation} 
\begin{aligned} 
&\text{A}_n = \text{softmax}(\boldsymbol{qk}^T \cdot \sqrt{d_{head}}) (h^{''}_{1:t}\boldsymbol{W}^n_V), \\ 
& \text{where} \ \ \boldsymbol{q} = \text{Norm}(h^{''}_t\boldsymbol{W}^n_Q) \odot s_{qk}, \\
& \quad \quad \quad \boldsymbol{k} = \text{Norm}(h^{''}_{1:t}\boldsymbol{W}^n_K) \odot s_{qk}, \\ 
\end{aligned} 
\end{equation}
with $s_{qk} \in \mathbb{R}^{d_{head}}$, and
\begin{equation} \begin{aligned} 
& \text{MLP}(h'_t) = (\text{SiLU}(v) \odot u)\boldsymbol{W}_{o\text{MLP}}, \\ 
& \text{where} \quad u = (h'_t\boldsymbol{W}_u) \odot s_u, \\ 
& \quad \quad \quad \ \ v = (h'_t\boldsymbol{W}_v) \odot s_v \sqrt{d_{\text{MLP}}}, 
\end{aligned} 
\end{equation}
with $s_u, s_v \in \mathbb{R}^{d_{\text{MLP}}}.$ Here, $s_{qk}, s_u, s_v$ are learnable scaling parameters introduced to compensate for the absence of the learnable affine parameters typically found in RMSNorm. Furthermore, all weight matrices $\boldsymbol{W}$ are normalized along their embedding dimension at every training step.

\section{Dataset Collection}
\begin{table*}[htp]
    \centering
    \begin{tabularx}{\textwidth}{l c >{\raggedright\arraybackslash}X} \toprule
        \textbf{Dataset} & \textbf{Train/Test} & \textbf{Labels} \\ \midrule
        Emolit & 6000/1200 & love, joy, excitement, surprise, anger, fear, disgust, sadness, boredom, calmness, relief, trust \\ \midrule
        Empathetic Dialogue & 4000/800 & \uline{joyful}, \uline{excited}, \uline{surprised}, \uline{angry}, \uline{afraid}, \uline{disgusted}, \uline{sad}, \uline{(trusting, faithful)} \\ \midrule
        SuperEmotion & 4050/900 & love, joy, excitement, surprise, anger, fear, disgust, sadness, relief \\ \midrule
        PersonaGen & 6000/1200 & love, joy, excitement, surprise, anger, fear, disgust, sadness, boredom, calmness, relief, trust \\ \bottomrule
    \end{tabularx}
    \caption{Details of the datasets used in Section~\ref{ssec: results}. All datasets are balanced, containing an equal number of samples for each emotion label. For Empathetic Dialogue, we chose the labels that most closely match the categories in Figure~\ref{fig:Emotion Wheel}.}
    \label{tab:dataset table}
\end{table*}
\begin{table*}[htbp]
\centering
\resizebox{\linewidth}{!}{%
\begin{tabular}{|l|c|c|c|c|c|c|}
\hline
\textbf{Pretrained Model} & 
  \makecell{intfloat/\\multilingual-e5-large} & 
  \makecell{mixedbread-ai/\\mxbai-embed-large-v1} & 
  \makecell{Qwen/\\Qwen3-Embedding-4B} & 
  \makecell{nvidia/\\llama-embed-nemotron-8b} & 
  \makecell{meta-llama/\\Llama-3.2-3B} & 
  \makecell{allenai/\\Olmo-3-1025-7B} \\
\hline
\textbf{Backbone}            & \multicolumn{2}{c|}{unfreeze} & \multicolumn{4}{c|}{freeze} \\
\hline
\textbf{Torch dtype}         & \multicolumn{6}{c|}{bfloat16} \\
\textbf{Epoch}               & \multicolumn{6}{c|}{15} \\
\textbf{Learning Rate}       & \multicolumn{6}{c|}{5e-5} \\
\textbf{LR scheduler type}   & \multicolumn{6}{c|}{constant} \\
\textbf{Train Batch Size}    & \multicolumn{6}{c|}{128} \\
\textbf{Random Seed}         & \multicolumn{6}{c|}{42} \\
\hline
\textbf{Hidden size $d$}     & \multicolumn{2}{c|}{1024} & 2560 & 4096 & 3072 & 4096 \\\hline
\textbf{Num attention heads} & \multicolumn{2}{c|}{16} & \multicolumn{2}{c|}{32} & 24 & 32 \\\hline
\textbf{Pooling strategy}    & \multicolumn{2}{c|}{cls} & last & mean & \multicolumn{2}{c|}{last} \\
\hline
\textbf{(SINCERE, SoftCSE)-$\tau$}         & \multicolumn{6}{c|}{0.05} \\\hline
\textbf{CircularCSE-$\text{margin} \ m$}         & \multicolumn{6}{c|}{0} \\\hline
\end{tabular}%
}
\caption{Hyperparameters used to train models}
\label{tab: hypra_backbone}
\end{table*}

\label{sec: dataset collection}
This section outlines the selection criteria and construction procedures for the dataset.

\subsection{Real-world Emotion Dataset} 
Our primary criterion for dataset selection was the availability of a diverse range of emotion labels. Reproducing a circular structure requires gathering emotions with distinct properties; in particular, low-arousal emotions (e.g., boredom, calmness) were critical yet often scarce in standard datasets.

\textbf{Emolit}~\cite{app13137502} is a large-scale emotion text dataset sourced from Project Gutenberg, annotated with 38 distinct emotion labels. Each text entry is accompanied by emotion prediction probabilities generated by a binary Natural Language Inference (NLI) model. The dataset is characterized by a well-balanced distribution of diverse emotion labels, achieved through rigorous preprocessing for noise removal and diversity enhancement. For our experiments, we assigned the emotion with the highest predicted probability as the ground truth label for each text.

\textbf{Empathetic Dialogue}~\cite{rashkin2019towards} is a conversational emotion dataset. Each session consists of a Speaker and a Listener, where the Listener's responses are designed to empathize with the Speaker's emotional state. Although the dataset is annotated with 32 emotion labels, it is not grounded in a specific psychological emotion model; consequently, it contains numerous labels that are undefined in standard psychological frameworks. In our preprocessing, we used the speaker's first utterance, and mapped 'afraid' to 'fear' and included 'faithful' under the 'trust' category to compensate for the deficiency of samples for the latter.

\textbf{SuperEmotion}~\cite{de2025super} is a unified dataset merging existing sources into Shaver’s six basic emotions and Neutral (seven classes). We employed subsets of GoEmotions~\cite{demszky2020goemotions} and CrowdFlower~\cite{van2012designing} from this collection. Although GoEmotions is highly popular, it is characterized by significant class imbalance, particularly regarding the scarcity of negative samples.

\subsection{Synthetic Emotion Dataset}
Due to the scarcity of datasets fully covering the circumplex emotion model, we employ an existing dataset synthesis framework to construct a new dataset specifically for our experiments. This approach mitigates the risk of representation space distortion caused by missing labels while ensuring a diverse range of emotional texts.

\textbf{PersonaGen}~\cite{inoshita2025persona} is a framework for generating diverse, high-fidelity text. It assigns persona attributes using census-based statistical distributions and utilizes LLMs to validate the consistency of these attribute combinations. This process ensures the creation of realistic personas that are highly likely to exist in reality.

\begin{figure}[htbp]
  \centering
\vspace{-10pt}
\begin{mybox}[PersonaGen prompt template]
\vspace{-9pt}
\raggedright  
\lstset{basicstyle=\scriptsize\ttfamily}
\begin{lstlisting}
###System prompt:### You are a roleplay AI.
###User prompt:### Roleplay as the persona 
below.
Speak 1-2 natural English sentences expressing 
the emotion.

[Persona] {'Age', 'Job', 'Education', 
           'Location', 'Family'}
[Scene] {'Scene'}
[Style] {'Style'}
[Emotion] {'Emotion'}
Output:
\end{lstlisting}
\vspace{-6pt}
\end{mybox}
\vspace{-9pt}
    \caption{Prompt template used in the dataset synthesis.}
  \label{fig:personagen template}
\end{figure}
Figure~\ref{fig:personagen template} shows the PersonaGen prompt template. We employed seven distinct LLMs to guarantee diversity in the synthesized dataset: 
\begin{itemize}
    \setlength{\itemsep}{0pt}
    \setlength{\parskip}{0pt}
    
    \small
    
    \item \texttt{gpt-5-nano}~\cite{gpt5nano}
    \item \texttt{Ministral-3-14B-Instruct-2512}~\cite{ministral3_14b}
    \item \texttt{Olmo-3-7B-Instruct}~\cite{olmo3_7b_instruct}
    \item \texttt{Qwen3-30B-A3B-Instruct-2507}~\cite{qwen3_30b_a3b_instruct_2507}
    \item \texttt{Phi-4-mini-instruct}~\cite{phi4_mini_instruct}
    \item \texttt{gemma-3-27b-it}~\cite{google2025gemma3}
    \item \texttt{Llama-3.3-70B-Instruct}~\cite{llama3_3_70b_instruct}
\end{itemize}
After randomly sampling from the model outputs, we removed duplicates and filtered for a length range of 3–50 tokens, ultimately collecting 600 texts for each emotion category.

\subsection{Integration with ECM}
The final dataset specifications are presented in Table~\ref{tab:dataset table}. Emotion labels for each dataset were filtered based on the configuration shown in Figure~\ref{fig:Emotion Wheel}. We manually selected emotions that either exhibited exact name matches or possessed equivalent semantic properties. Furthermore, we balanced the emotion label distribution in each dataset through under-sampling to avoid embedding space distortion arising from class imbalance. The findings in Section~\ref{ssec: results} are based on these datasets.
\subsection{Differences between the ECM and Russell's Circumplex Model of Affect}
\label{ssec: label difference}
Notable deviations from Russell's Circumplex Model of Affect include the placement of love at 0°, calmness at 270°, and trust at 330°.
\begin{itemize}
    \item 0° (love): While this angle typically represents pleasant emotions such as "satisfied" or "pleased," we adopted "love" because it is a predominant emotion in real-world datasets and is considered the polar opposite of "disgust."
    \item 270° (calmness): This angle normally corresponds to "tired" or "sleepy." However, to the best of our knowledge, no text datasets annotated with these specific labels exist. Consequently, we substituted them with "calmness," which serves as the closest proxy for a deactivated emotional state.
    \item 330° (trust): While "serene" is the standard representative for this angle, it is rare within corpora. We therefore adopted "trust" as it approximates the polar opposite of "fear."
\end{itemize}
As discussed in the \textit{Limitations} section, since our experimental results are primarily driven by dimensional constraints, the specific arrangement or selection of emotion types does not significantly alter our conclusions. However, effectively capturing rare emotions within training corpora remains a challenge for future work. We posit that CircularCSE, with its ability to model inter-emotional relationships, is particularly advantageous for learning low-frequency emotion labels.
\section{Implementation Details}
\label{sec: implementation details}
Table~\ref{tab: hypra_backbone} lists the hyperparameters used for training. These values were determined through preliminary experiments involving various combinations, selected to ensure training stability. For any parameters not explicitly listed, we employ the default settings.
\section{Overall Results}
Table~\ref{tab:performance_main_extended} presents the results obtained by training six distinct backbone models across each dataset, applying six combinations of head architectures and loss functions. Observing the trends across datasets, we find that all methods consistently demonstrate significantly high accuracy on PersonaGen, whereas performance remains low on Empathetic Dialogue and SuperEmotion.

This suggests that compared to real-world data, synthetic data lacks diversity and ambiguity, making it an inherently easier dataset for emotion prediction. Conversely, Empathetic Dialogue and SuperEmotion originate from conversations and social media, respectively; these results imply that accurate emotion prediction on such datasets is difficult without clear context.
\label{sec: overall results}
\begin{table*}[t]
\centering
\scriptsize
\setlength{\tabcolsep}{3pt}
\begin{tabular}{llcccccccc|cc}
\toprule
\multirow{2}{*}{Base Model} & \multirow{2}{*}{Method} &
\multicolumn{2}{c}{Emolit} &
\multicolumn{2}{c}{Empathetic Dialogue} &
\multicolumn{2}{c}{SuperEmotion} &
\multicolumn{2}{c|}{PersonaGen} &
\multicolumn{2}{c}{Average} \\
\cmidrule(lr){3-12}
& & $V_{\text{Measure}}$ & CD-r & $V_{\text{Measure}}$ & CD-r & $V_{\text{Measure}}$ & CD-r & $V_{\text{Measure}}$ & CD-r & $V_{\text{Measure}}$ & CD-r \\
\midrule

\multirow{9}{*}{\shortstack[l]{mE5}} 
& Pretrained & 0.437 & 0.631 & 0.237 & 0.512 & 0.170 & 0.554 & 0.525 & 0.599 & 0.342 & 0.574 \\
\cmidrule(lr){2-12}\morecmidrules\cmidrule(lr){2-12}
& \multicolumn{11}{l}{\textbf{SINCERE}} \\
& \hspace{0.5em}- GPT & \textbf{0.835} & 0.402 & \textbf{0.667} & \uline{0.413} & 0.621 & \uline{0.343} & 0.917 & \uline{0.112} & \textbf{0.760} & 0.317 \\
& \hspace{0.5em}- nGPT & 0.822 & \uline{0.068} & 0.643 & 0.504 & 0.595 & 0.186 & 0.916 & 0.125 & 0.744 & \uline{0.221} \\
\cmidrule(lr){2-12}
& \multicolumn{11}{l}{\textbf{SoftCSE}} \\
& \hspace{0.5em}- GPT & 0.817 & 0.519 & 0.649 & 0.514 & \textbf{0.636} & 0.574 & 0.915 & 0.302 & 0.755 & 0.477 \\
& \hspace{0.5em}- nGPT & 0.822 & 0.468 & 0.660 & \textbf{0.575} & 0.621 & 0.559 & \uline{0.909} & 0.395 & 0.753 & 0.499 \\
\cmidrule(lr){2-12}
& \multicolumn{11}{l}{\textbf{CircularCSE}} \\
& \hspace{0.5em}- GPT & 0.796 & 0.894 & \uline{0.593} & 0.551 & 0.570 & 0.688 & \uline{0.909} & 0.896 & \uline{0.717} & 0.757 \\
& \hspace{0.5em}- nGPT & \uline{0.787} & \textbf{0.904} & 0.610 & 0.537 & \uline{0.565} & \textbf{0.710} & \textbf{0.918} & \textbf{0.907} & 0.720 & \textbf{0.764} \\
\midrule
\midrule

\multirow{9}{*}{\shortstack[l]{mxbai}} 
& Pretrained & 0.603 & 0.740 & 0.502 & 0.326 & 0.285 & 0.490 & 0.670 & 0.625 & 0.515 & 0.545 \\
\cmidrule(lr){2-12}\morecmidrules\cmidrule(lr){2-12}
& \multicolumn{11}{l}{\textbf{SINCERE}} \\
& \hspace{0.5em}- GPT & 0.831 & 0.293 & 0.640 & \uline{0.408} & 0.609 & \uline{0.212} & 0.909 & 0.234 & 0.747 & \uline{0.287} \\
& \hspace{0.5em}- nGPT & 0.825 & \uline{0.159} & 0.648 & 0.492 & 0.595 & 0.482 & 0.912 & \uline{0.226} & 0.745 & 0.340 \\
\cmidrule(lr){2-12}
& \multicolumn{11}{l}{\textbf{SoftCSE}} \\
& \hspace{0.5em}- GPT & \textbf{0.841} & 0.645 & \textbf{0.657} & 0.584 & 0.595 & 0.479 & 0.912 & 0.686 & \textbf{0.751} & 0.599 \\
& \hspace{0.5em}- nGPT & 0.826 & 0.577 & 0.628 & \textbf{0.636} & \textbf{0.610} & 0.506 & \textbf{0.927} & 0.619 & 0.748 & 0.585 \\
\cmidrule(lr){2-12}
& \multicolumn{11}{l}{\textbf{CircularCSE}} \\
& \hspace{0.5em}- GPT & 0.840 & \textbf{0.893} & 0.623 & 0.513 & \uline{0.568} & 0.700 & 0.912 & 0.899 & 0.736 & 0.751 \\
& \hspace{0.5em}- nGPT & \uline{0.793} & 0.886 & \uline{0.604} & 0.520 & 0.572 & \textbf{0.705} & \uline{0.907} & \textbf{0.910} & \uline{0.719} & \textbf{0.755} \\
\midrule
\midrule

\multirow{9}{*}{\shortstack[l]{Qwen3\\-Embedding\\-4B}} 
& Pretrained & 0.579 & 0.556 & 0.435 & 0.442 & 0.273 & 0.440 & 0.691 & 0.651 & 0.495 & 0.522 \\
\cmidrule(lr){2-12}\morecmidrules\cmidrule(lr){2-12}
& \multicolumn{11}{l}{\textbf{SINCERE}} \\
& \hspace{0.5em}- GPT & \textbf{0.859} & \uline{0.122} & 0.651 & 0.595 & \textbf{0.588} & 0.490 & 0.927 & \uline{0.013} & \textbf{0.756} & \uline{0.305} \\
& \hspace{0.5em}- nGPT & 0.840 & 0.519 & \textbf{0.673} & \uline{0.471} & 0.554 & 0.579 & 0.895 & 0.613 & 0.741 & 0.545 \\
\cmidrule(lr){2-12}
& \multicolumn{11}{l}{\textbf{SoftCSE}} \\
& \hspace{0.5em}- GPT & 0.846 & 0.647 & 0.646 & \textbf{0.673} & 0.578 & \uline{0.487} & \textbf{0.932} & 0.403 & 0.751 & 0.552 \\
& \hspace{0.5em}- nGPT & 0.836 & 0.729 & 0.643 & 0.587 & 0.523 & \textbf{0.731} & 0.890 & 0.786 & 0.723 & 0.708 \\
\cmidrule(lr){2-12}
& \multicolumn{11}{l}{\textbf{CircularCSE}} \\
& \hspace{0.5em}- GPT & \uline{0.709} & \textbf{0.893} & \uline{0.530} & 0.520 & 0.471 & 0.675 & 0.860 & 0.900 & \uline{0.643} & 0.747 \\
& \hspace{0.5em}- nGPT & 0.741 & 0.880 & 0.603 & 0.560 & \uline{0.443} & 0.660 & \uline{0.850} & \textbf{0.911} & 0.659 & \textbf{0.753} \\
\midrule
\midrule

\multirow{9}{*}{\shortstack[l]{Llama\\-Embed\\-Nemotron\\-8B}} 
& Pretrained & 0.271 & 0.373 & 0.227 & 0.258 & 0.160 & 0.118 & 0.494 & 0.715 & 0.288 & 0.366 \\
\cmidrule(lr){2-12}\morecmidrules\cmidrule(lr){2-12}
& \multicolumn{11}{l}{\textbf{SINCERE}} \\
& \hspace{0.5em}- GPT & 0.841 & \uline{-0.041} & \textbf{0.696} & 0.591 & 0.601 & 0.461 & 0.940 & \uline{-0.293} & 0.769 & \uline{0.180} \\
& \hspace{0.5em}- nGPT & 0.834 & 0.386 & 0.693 & \uline{0.345} & \textbf{0.627} & \uline{0.393} & 0.938 & 0.617 & \textbf{0.773} & 0.435 \\
\cmidrule(lr){2-12}
& \multicolumn{11}{l}{\textbf{SoftCSE}} \\
& \hspace{0.5em}- GPT & \textbf{0.847} & 0.577 & 0.685 & \textbf{0.691} & 0.600 & 0.423 & \textbf{0.941} & 0.339 & 0.768 & 0.508 \\
& \hspace{0.5em}- nGPT & \textbf{0.847} & 0.735 & 0.692 & 0.591 & 0.599 & 0.657 & 0.938 & 0.756 & 0.769 & 0.685 \\
\cmidrule(lr){2-12}
& \multicolumn{11}{l}{\textbf{CircularCSE}} \\
& \hspace{0.5em}- GPT & 0.710 & \textbf{0.902} & 0.603 & 0.508 & 0.519 & \textbf{0.675} & 0.903 & 0.908 & 0.684 & \textbf{0.748} \\
& \hspace{0.5em}- nGPT & \uline{0.685} & 0.901 & \uline{0.605} & 0.483 & \uline{0.447} & 0.671 & \uline{0.880} & \textbf{0.913} & \uline{0.654} & 0.742 \\

\midrule
\midrule

\multirow{9}{*}{\shortstack[l]{Llama-3.2\\-3B}} 
& Pretrained & 0.100 & 0.064 & 0.069 & 0.067 & 0.050 & 0.101 & 0.159 & 0.636 & 0.094 & 0.217 \\
\cmidrule(lr){2-12}\morecmidrules\cmidrule(lr){2-12}
& \multicolumn{11}{l}{\textbf{SINCERE}} \\
& \hspace{0.5em}- GPT & \textbf{0.780} & \uline{0.404} & \textbf{0.627} & 0.548 & \textbf{0.566} & \uline{0.363} & \textbf{0.926} & \uline{0.116} & \textbf{0.725} & \uline{0.358} \\
& \hspace{0.5em}- nGPT & 0.752 & 0.419 & 0.357 & \uline{0.262} & 0.296 & 0.566 & 0.903 & 0.451 & 0.577 & 0.425 \\
\cmidrule(lr){2-12}
& \multicolumn{11}{l}{\textbf{SoftCSE}} \\
& \hspace{0.5em}- GPT & 0.760 & 0.600 & 0.616 & \textbf{0.723} & 0.549 & 0.476 & 0.913 & 0.391 & 0.710 & 0.548 \\
& \hspace{0.5em}- nGPT & 0.628 & 0.881 & 0.362 & 0.522 & 0.276 & 0.676 & 0.798 & 0.833 & 0.516 & \textbf{0.728} \\
\cmidrule(lr){2-12}
& \multicolumn{11}{l}{\textbf{CircularCSE}} \\
& \hspace{0.5em}- GPT & 0.588 & 0.891 & 0.484 & 0.457 & 0.425 & 0.657 & 0.817 & \textbf{0.909} & 0.579 & \textbf{0.728} \\
& \hspace{0.5em}- nGPT & \uline{0.489} & \textbf{0.902} & \uline{0.250} & 0.465 & \uline{0.196} & 0.564 & \uline{0.592} & 0.900 & \uline{0.382} & 0.708 \\
\midrule
\midrule

\multirow{9}{*}{\shortstack[l]{Olmo-3\\-7B}} 
& Pretrained & 0.047 & -0.005 & 0.024 & -0.036 & 0.048 & -0.014 & 0.133 & 0.265 & 0.063 & 0.053 \\
\cmidrule(lr){2-12}\morecmidrules\cmidrule(lr){2-12}
& \multicolumn{11}{l}{\textbf{SINCERE}} \\
& \hspace{0.5em}- GPT & \textbf{0.805} & \uline{0.424} & \textbf{0.615} & 0.485 & \textbf{0.581} & \uline{0.403} & 0.929 & \uline{0.169} & \textbf{0.732} & 0.370 \\
& \hspace{0.5em}- nGPT & 0.786 & 0.444 & 0.409 & \uline{0.297} & 0.376 & 0.467 & 0.913 & 0.243 & 0.621 & \uline{0.363} \\
\cmidrule(lr){2-12}
& \multicolumn{11}{l}{\textbf{SoftCSE}} \\
& \hspace{0.5em}- GPT & \textbf{0.805} & 0.578 & 0.609 & \textbf{0.579} & 0.561 & 0.425 & \textbf{0.941} & 0.380 & 0.729 & 0.490 \\
& \hspace{0.5em}- nGPT & 0.742 & 0.800 & 0.351 & 0.546 & 0.335 & 0.636 & 0.898 & 0.701 & 0.582 & 0.671 \\
\cmidrule(lr){2-12}
& \multicolumn{11}{l}{\textbf{CircularCSE}} \\
& \hspace{0.5em}- GPT & 0.630 & \textbf{0.903} & 0.484 & 0.487 & 0.414 & \textbf{0.682} & 0.844 & 0.902 & 0.593 & \textbf{0.744} \\
& \hspace{0.5em}- nGPT & \uline{0.494} & 0.876 & \uline{0.248} & 0.418 & \uline{0.254} & 0.533 & \uline{0.605} & \textbf{0.914} & \uline{0.400} & 0.685 \\

\bottomrule
\end{tabular}
\caption{V-Measure and Circumplex Distance correlation (CD-r) across datasets and models. \textbf{Bold} indicates the maximum value and \underline{underlined} indicates the minimum value across different configurations for each model.}
\label{tab:performance_main_extended}
\end{table*}

\section{Robustness Experiments}
\label{sec: robustness detail}
\begin{figure*}[!tp]
  \centering
  \begin{subfigure}{0.32\linewidth}
    \includegraphics[width=\linewidth]{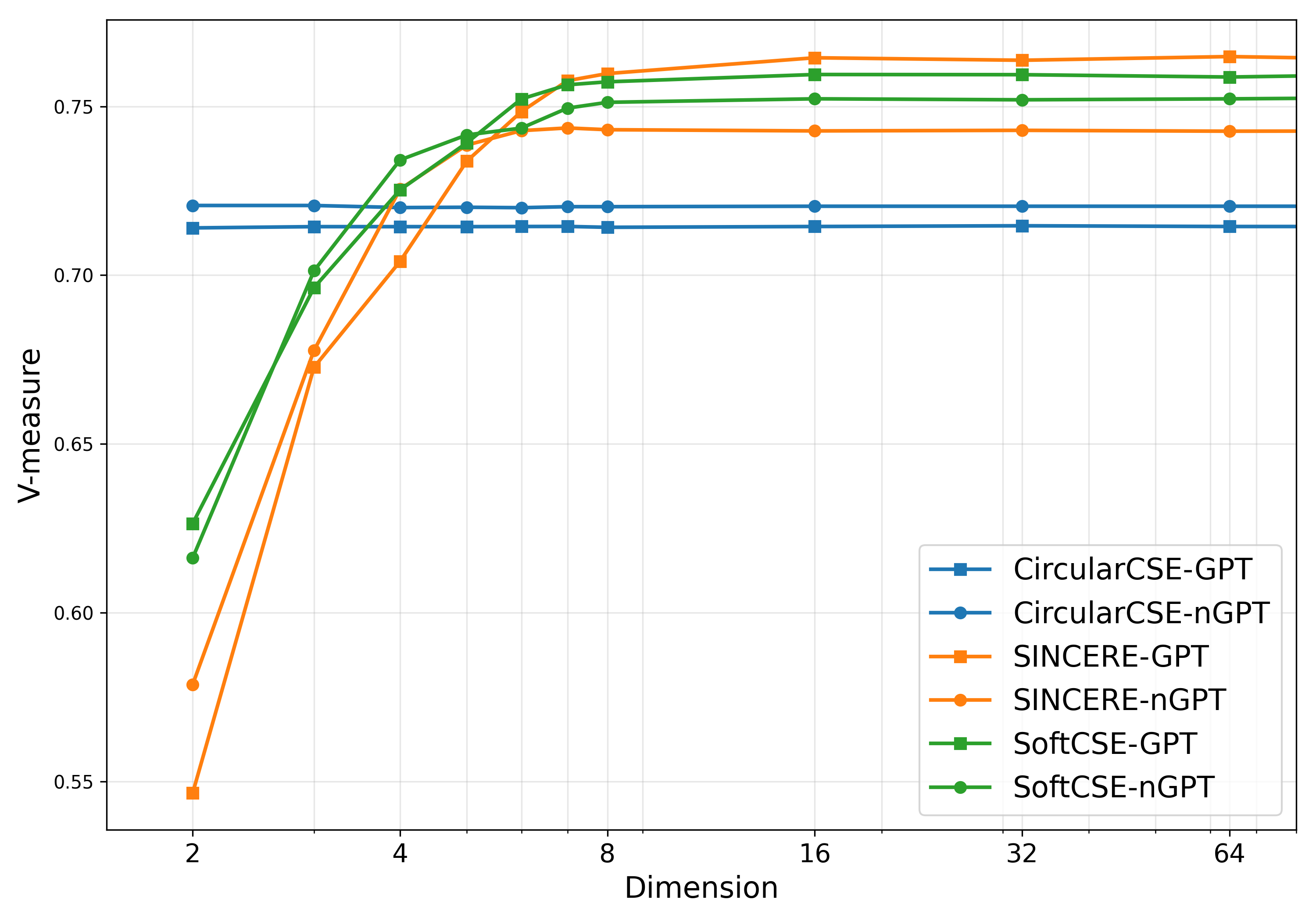}
    \caption{mE5}
  \end{subfigure}
  \hfill
  \begin{subfigure}{0.32\linewidth}
    \includegraphics[width=\linewidth]{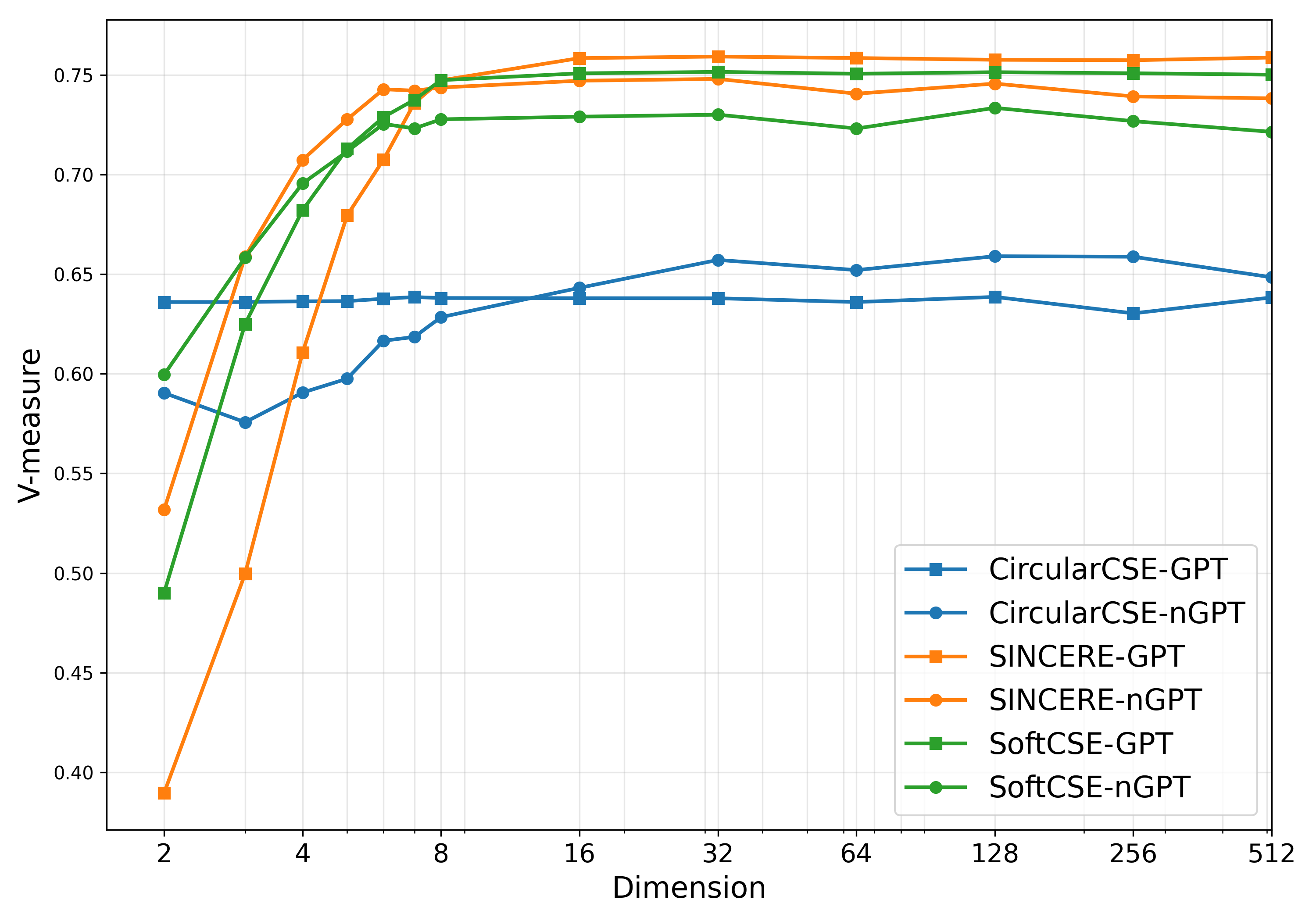}
    \caption{Qwen3-Embedding-4B}
  \end{subfigure}
  \hfill
  \begin{subfigure}{0.32\linewidth}
    \includegraphics[width=\linewidth]{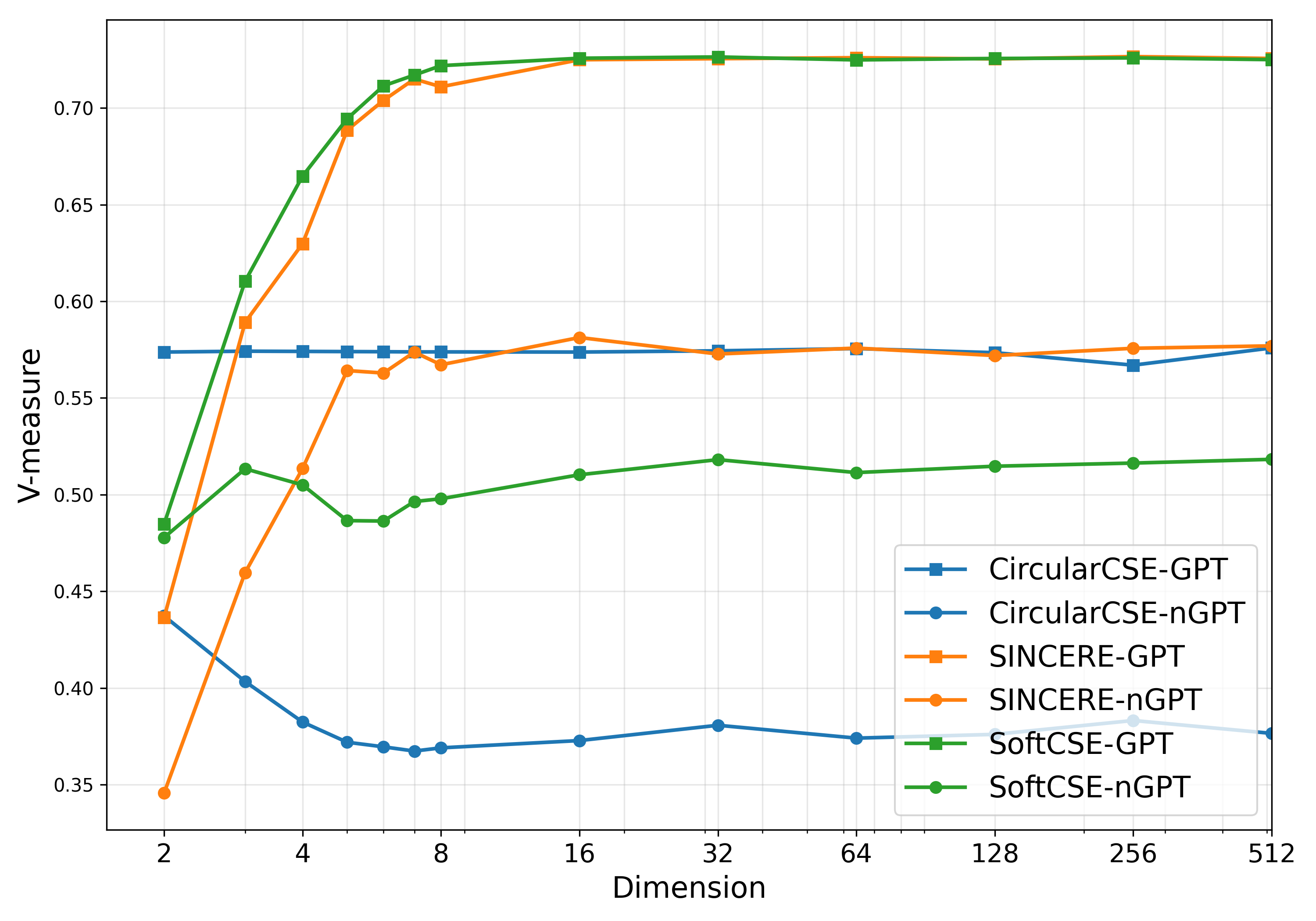}
    \caption{Llama-3.2-3B}
  \end{subfigure}
  \hfill
  \caption{Robustness to dimensionality reduction.}
  \label{fig: dimensionality reduction extended}
\end{figure*}

\begin{figure*}[!tp]
    \centering
    \includegraphics[width=0.8\linewidth]{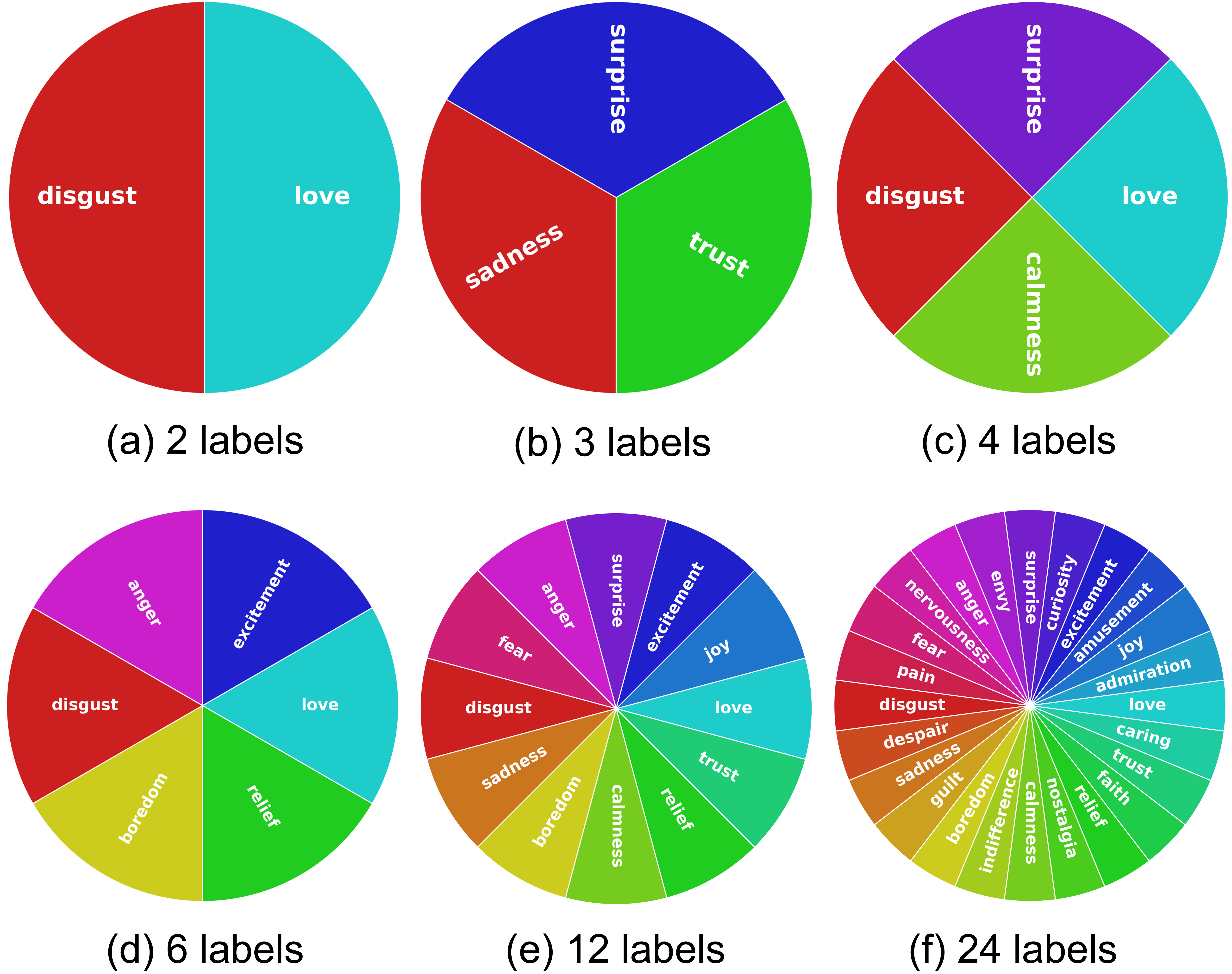}
    \caption{Empirical circumplex models used in label variation experiments}
    \label{fig:EWD_ablation}
\end{figure*}

\begin{figure*}[!tp]
  \centering
  \begin{subfigure}{0.32\linewidth}
    \includegraphics[width=\linewidth]{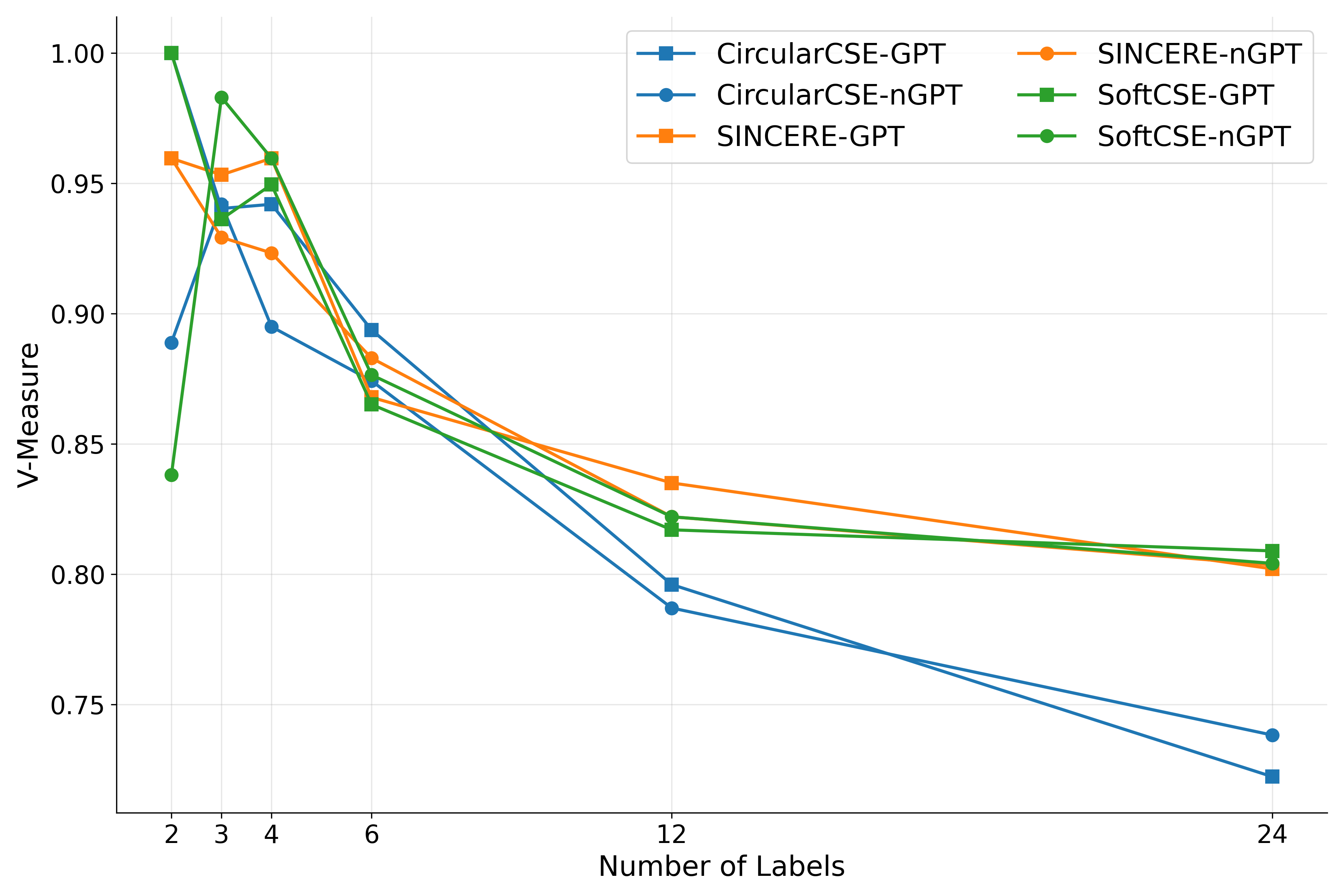}
    \caption{mE5}
  \end{subfigure}
  \hfill
  \begin{subfigure}{0.32\linewidth}
    \includegraphics[width=\linewidth]{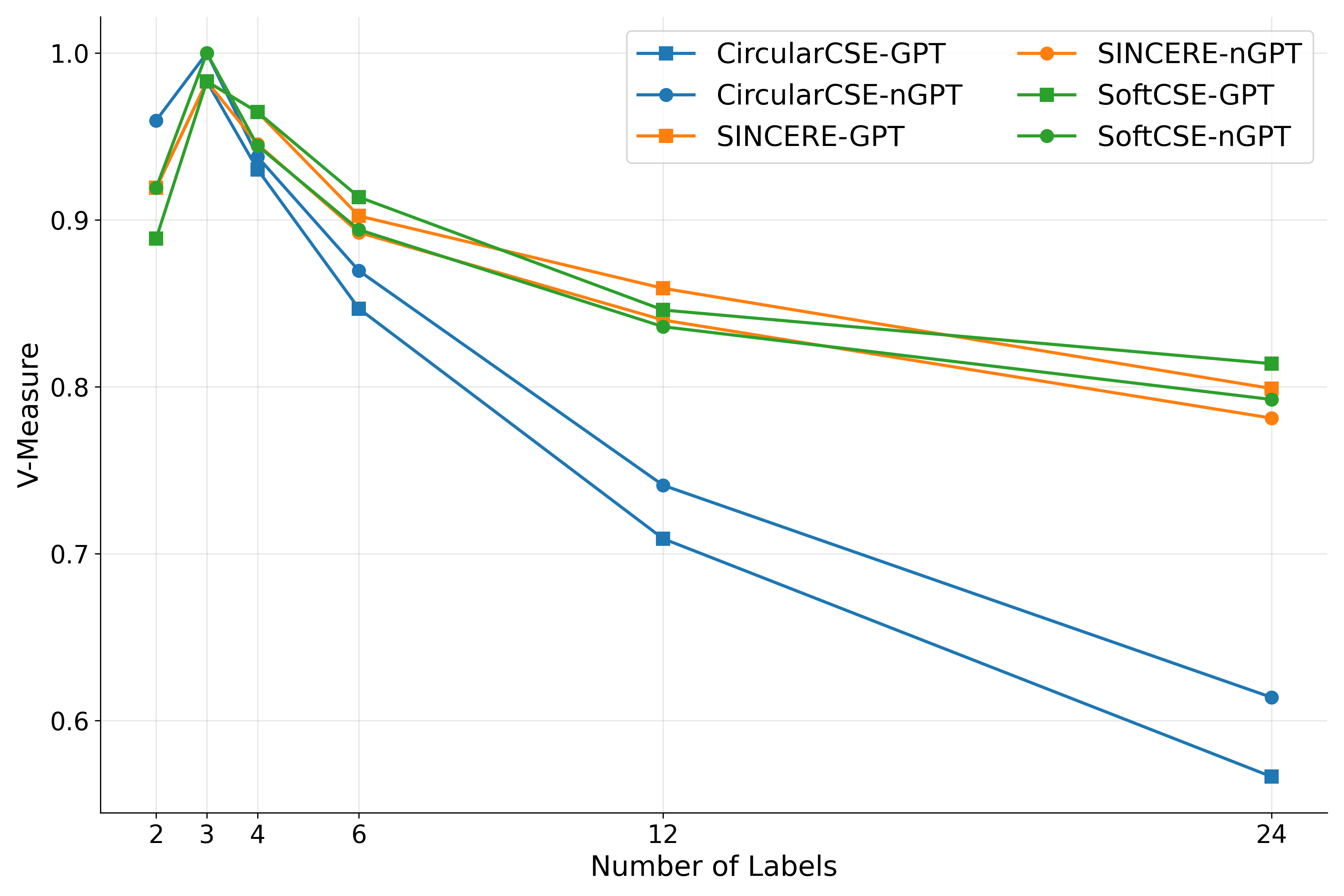}
    \caption{Qwen3-Embedding-4B}
  \end{subfigure}
  \hfill
  \begin{subfigure}{0.32\linewidth}
    \includegraphics[width=\linewidth]{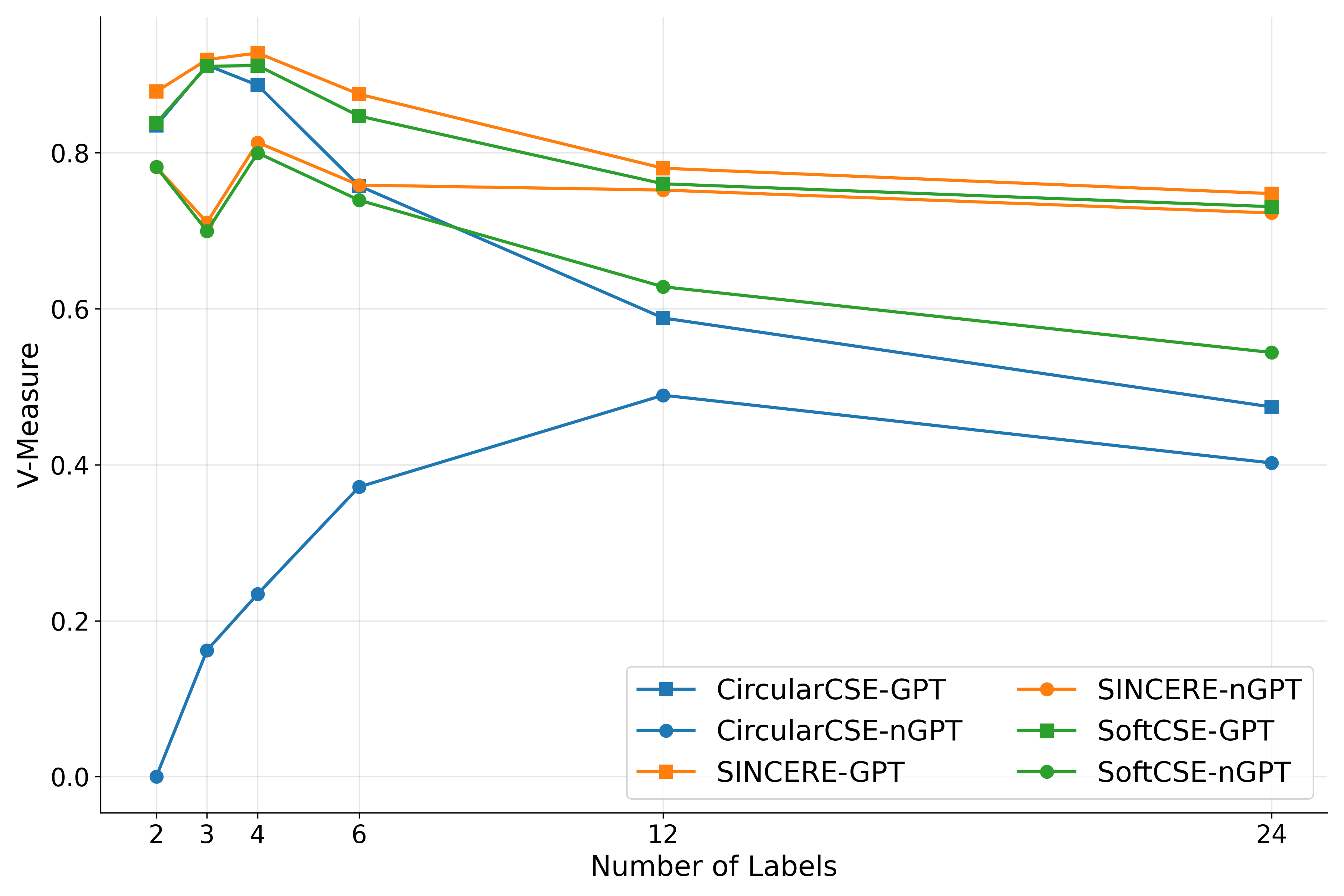}
    \caption{Llama-3.2-3B}
  \end{subfigure}
  \hfill
  \caption{Robustness to the number of emotion labels.}
  \label{fig: label ablation extended}
\end{figure*}

We illustrate the dimensionality robustness of mE5, Qwen3-Embedding-4B, and Llama-3.2-3B in Figure~\ref{fig: dimensionality reduction extended}, and the robustness to label count variations in Figure~\ref{fig: label ablation extended}. Experiments regarding variations in the number of labels were conducted using the Emolit dataset. Figure~\ref{fig:EWD_ablation} illustrates the arrangement and types of emotions within the circular model. We manually determined these label placements based on Russell's Circumplex Model~\cite{Russell1980, 12-point_circumplex}.

\section{Lower Bound of SINCERE}
\label{sec: loss function math}
In this section, we provide a simplified proof regarding the lower bound of SINCERE. To simplify the derivation, we posit the following assumption.

\noindent$\textbf{Assumption 1} \ (\text{Class Prototype}).$

\noindent \textit{The representation of each class $y_i \in \mathcal{E}=\{y_1, \dots, y_E\}$ is concentrated at a single point, denoted by the unit vector $e_i \in \mathbb{S}^{d-1}, \left\|e_i\right\|=1$, and the positive example corresponding to an anchor is always $e_i$.}

\noindent$\textbf{Assumption 2} \ (\text{Class Uniformity}).$

\noindent \textit{The dataset is class-balanced, i.e., 
$p(y=i)=\frac{1}{E}.$ We analyze the population (expected) SINCERE objective, under which an anchor from class $y_i$ contrasts against the representations of all other classes in expectation. Consequently, the negative set consists of the 
$E - 1 $ class prototypes $\{e_j\}_{j \neq i}$.}

\noindent Under these assumptions, given the formulation:
{
\small
\begin{equation}
\begin{aligned}
& \mathcal{L}_{\mathrm{SINCERE}} = \\
&\mathbb{E}_{i}\mathbb{E}_{j \sim \mathcal{P}}\biggl\lbrack-\log
\frac{\exp(e^T_ie_j / \tau)}{\exp(e^T_ie_j / \tau) + \sum_{k \in \mathcal{N}}\exp(e^T_ie_k / \tau)}\biggr\rbrack
\end{aligned}
\end{equation}
}
the following holds:

\noindent$\textbf{Theorem}.$

\noindent \textit{When there are $E$ emotion classes, the lower bound of $\mathcal{L}_{\text{SINCERE}}$ is achieved when the class representations form a regular simplex, resulting in a positive-negative inner product (similarity) of $-\frac{1}{E-1}$. }

\noindent Here, a regular simplex is a configuration where the class representations satisfy the following conditions: 
\begin{itemize}
    \item Zero Sum: $\sum_{i=1}^E e_i = 0$
    \item Equal Norms 
    \item Equal Pairwise Inner Products
\end{itemize}
 It is well-established that Cross Entropy Loss and Supervised Contrastive Loss also converge to a regular simplex under certain conditions~\cite{doi:10.1073/pnas.2015509117, pmlr-v139-graf21a}. For a more rigorous proof, please refer to~\cite{doi:10.1073/pnas.2015509117, pmlr-v139-graf21a}.

\noindent \textit{Proof.} 
{
\small
\begin{equation}
\begin{aligned}
& \mathcal{L}_{\mathrm{SINCERE}} = \\
&\mathbb{E}_{i}\mathbb{E}_{j \sim \mathcal{P}}\biggl\lbrack
-\log
\frac{\exp(e^T_ie_j / \tau)}{\exp(e^T_ie_j / \tau) + \sum_{k \in \mathcal{N}}\exp(e^T_ie_k / \tau)}
\biggr\rbrack \\
& = \mathbb{E}_{i}\mathbb{E}_{j \sim \mathcal{P}}\biggl\lbrack
-\log
\frac{\exp(1 / \tau)}{\exp(1 / \tau) + \sum_{k \in \mathcal{N}}\exp(e^T_ie_k / \tau)}
\biggr\rbrack \\
&= \mathbb{E}_i\mathbb{E}_{j \sim \mathcal{P}}\left[ \log \left( 1 + \sum_{k \in \mathcal{N}} \exp\left(\frac{e_i^T e_k - 1}{\tau}\right) \right) \right] 
\end{aligned}
\end{equation}
}
Given that $\tau > 0$, $\mathcal{L}_{\text{SINCERE}}$ is monotonically increasing with respect to $\exp(e^T_i e_k / \tau)$. Consequently, minimizing $\mathcal{L}_{\text{SINCERE}}$ is equivalent to minimizing the negative sample term $\sum_{k \in \mathcal{N}} \exp(e^T_i e_k / \tau)$ for each sample.
For the set of class vectors $\{e_1, \dots, e_E\} \subset \mathbb{S}^{d-1}$, the squared $L^2$ norm of their sum is given by:
{
\small
\begin{equation}
\left\| \sum_{i=1}^E e_i \right\|^2 = \sum_{i=1}^E \|e_i\|^2 + \sum_{i \neq k} e^T_i e_k = E + \sum_{i \neq k} e^T_i e_k \geq 0
\end{equation}
}
Therefore, it follows that: $\sum_{i \neq k} e^T_i e_k \geq -E$.
Let $\overline{e^T_i e_k}$ denote the mean of the inner products for all pairs. Using the relationship $\sum_{i \neq k} e^T_i e_k = E(E-1)\overline{e^T_i e_k}$, we derive the inequality
\begin{equation}
\label{eq: average inner product}
\overline{e^T_i e_k} \geq -\frac{1}{E-1}
\end{equation}
This implies that the global average of the inner products between a class vector and vectors of other classes is lower-bounded by $-\frac{1}{E-1}$.
We consider the lower bound of the negative term. Since the exponential function is convex, applying Jensen's inequality:
{
\small
\begin{equation}
\sum_{k \in \mathcal{N}} \frac{1}{E-1} \exp\left(\frac{e^T_i e_k}{\tau}\right) \geq \exp\left( \sum_{k \in \mathcal{N}} \frac{1}{E-1} \frac{e^T_i e_k}{\tau} \right)
\end{equation}
}
Based on Assumption 2, we derive:
\begin{equation}    
\sum_{k \in \mathcal{N}} \exp\left(\frac{e^T_i e_k}{\tau}\right) \geq (E-1) \exp\left(\frac{\overline{e^T_i e_k}}{\tau}\right)
\end{equation}
Applying Jensen's inequality again, 
\begin{equation}
\begin{aligned}
&\sum_{i=1}^E \sum_{k \in \mathcal{N}} \frac{1}{E} \exp\left(\frac{e^T_i e_k}{\tau}\right) \\
&\geq (E-1) \sum_{i=1}^E \frac{1}{E} \exp\left(\frac{\overline{e^T_i e_k}}{\tau}\right) \\
& \geq (E-1) \exp\left(\sum_{i=1}^E \frac{1}{E} \frac{\overline{e^T_i e_k}}{\tau}\right)
\end{aligned}
\end{equation}
Based on Equation~(\ref{eq: average inner product}),
{
\small
\begin{equation}
\sum_{i=1}^E \sum_{k \in \mathcal{N}} \exp\left(\frac{e^T_i e_k}{\tau}\right) \geq E(E-1) \exp\left(\frac{-1}{(E-1)\tau}\right)
\end{equation}
}
This is the lower bound of SINCERE Loss based on the Assumption. The condition for equality in Jensen's inequality dictates that all pairwise inner products between class vectors must be equal, and their mean must be $-\frac{1}{E-1}$. Consequently, the class similarity between positive and negative samples required to achieve the lower bound is given by 
\begin{equation}  
e^T_i e_k = -\frac{1}{E-1} \  \forall\ i \neq k
\end{equation}
Furthermore, the condition for equality in Equation~(\ref{eq: average inner product}) is given by:
\begin{equation}
\left\| \sum_{i=1}^E e_i \right\|^2 = 0 \iff \sum_{i=1}^E e_i = 0
\end{equation}
Summarizing these conditions, it is shown that the geometry required to achieve the minimum loss satisfies the properties of a regular simplex.

\section{MDS Visualization}
\label{sec: module visualization}
\begin{figure*}
    \centering
    \includegraphics[width=\linewidth]{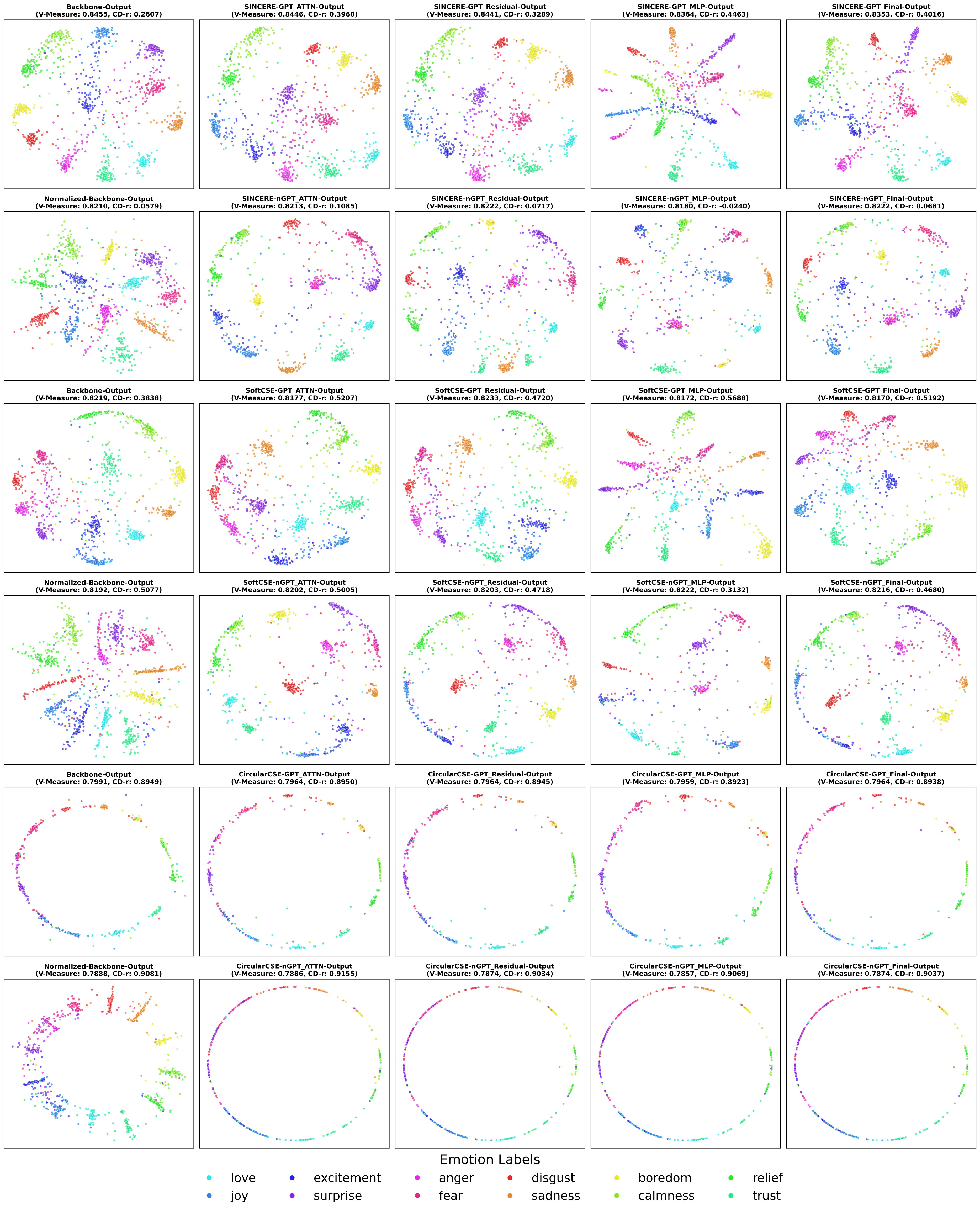}
    \caption{MDS visualization of mE5}
    \label{fig:mds_module_mE5}
\end{figure*}

\begin{figure*}
    \centering
    \includegraphics[width=\linewidth]{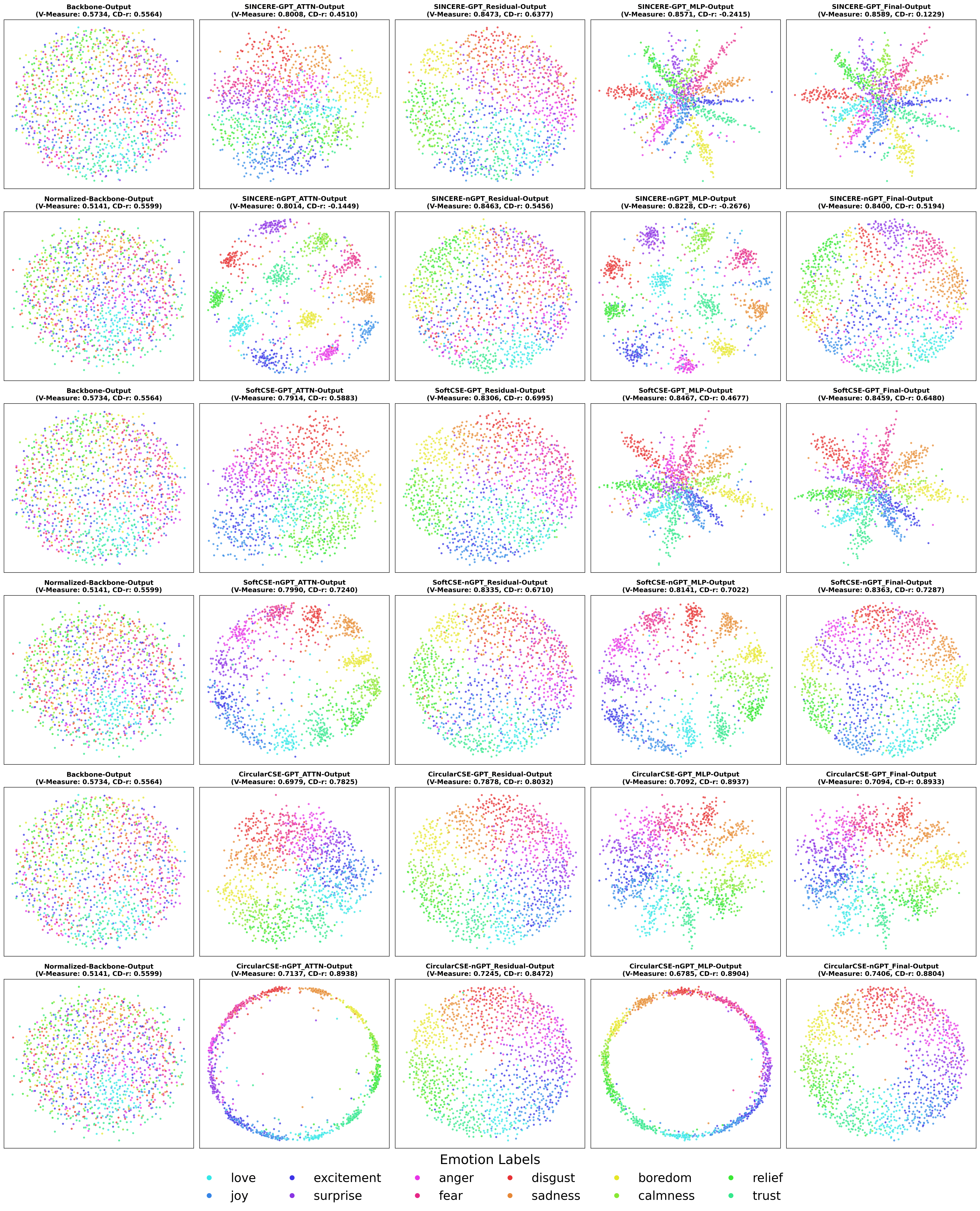}
    \caption{MDS visualization of Qwen3-Embedding-4B}
    \label{fig:mds_module_Qwen}
\end{figure*}

\begin{figure*}
    \centering
    \includegraphics[width=\linewidth]{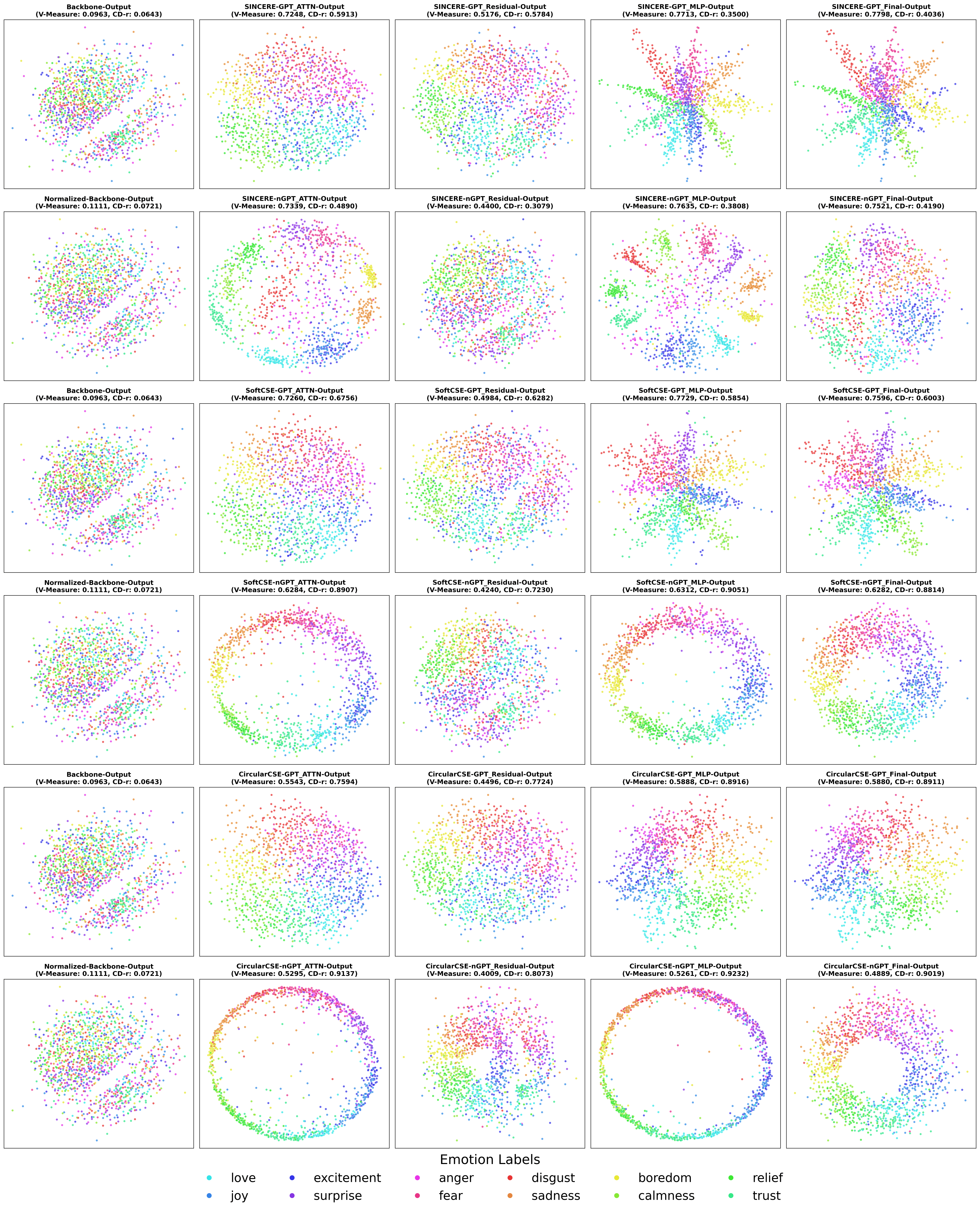}
    \caption{MDS visualization of Llama-3.2-3B}
    \label{fig:mds_module_Llama}
\end{figure*}

\begin{figure*}
    \centering
    \includegraphics[width=\linewidth]{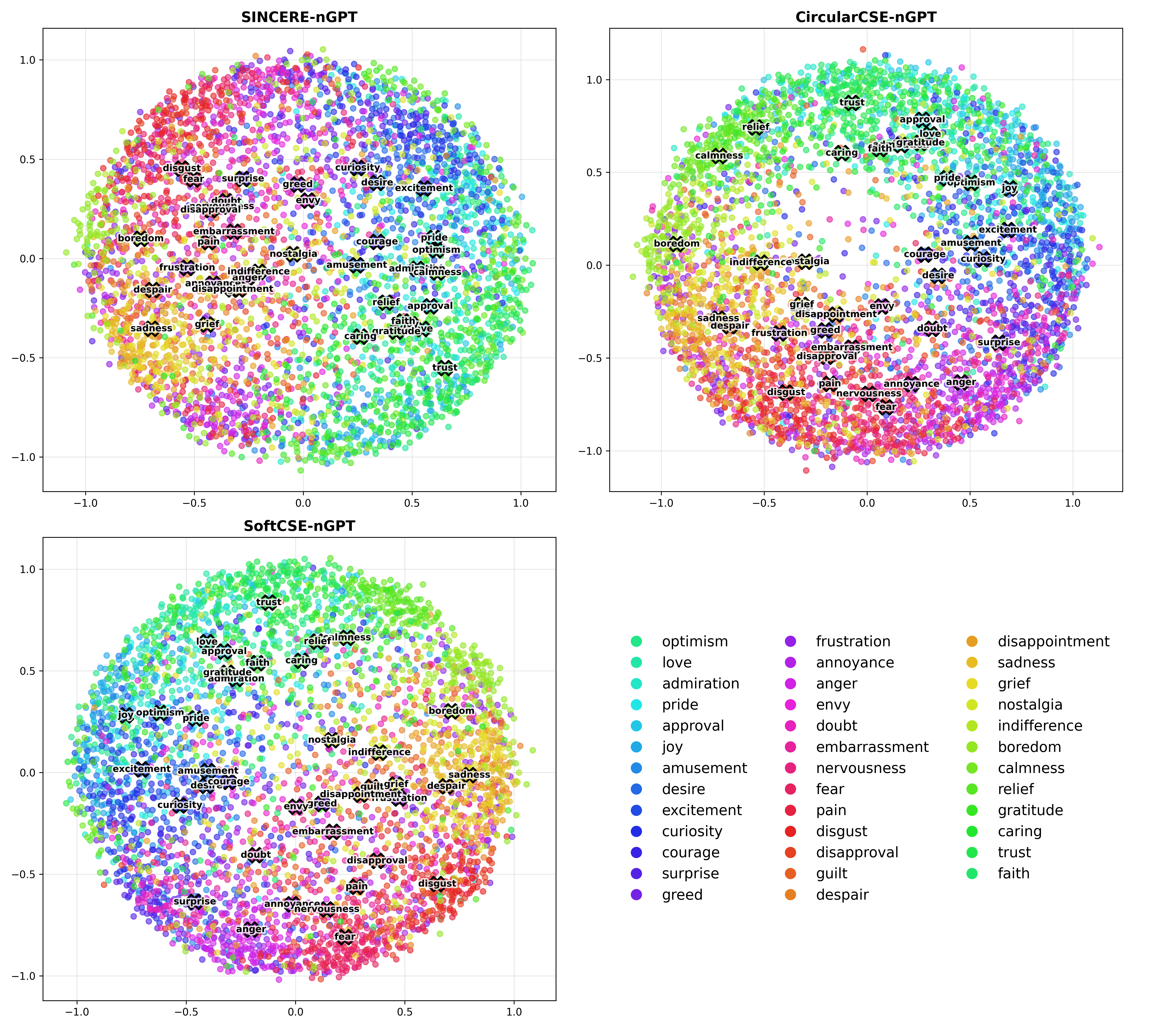}
    \caption{MDS plot of all emotion labels in the Emolit dataset for Qwen3-Embedding-4B trained with ECM. The plotted label names represent the centroids for each emotion class.}
    \label{fig:mds_all_label}
\end{figure*}

Figures~\ref{fig:mds_module_mE5}, ~\ref{fig:mds_module_Qwen}, and ~\ref{fig:mds_module_Llama} display the visualization results of the embedding representations from each module of mE5, Qwen3-Embedding-4B, Llama-3.2-3B on the Emolit dataset, projected using MDS. Figure~\ref{fig:mds_all_label} presents a MDS plot of the embeddings from the nGPT head of Qwen3-Embedding-4B, trained on the 12 emotion labels of Emolit used in Section~\ref{ssec: results}. This visualization encompasses all 38 emotion labels from the Emolit dataset, including those not seen during training. CircularCSE demonstrates the ability to position even unseen emotion labels in generally plausible locations. Furthermore, the region surrounding "boredom" and "calmness" is sparsely populated; this suggests that such visualizations can be utilized to examine the bias and comprehensiveness of emotion labels within a newly proposed dataset.

\end{document}